\documentclass[10pt,twocolumn,letterpaper]{article}

\usepackage{iccv}
\usepackage{times}
\usepackage{epsfig}
\usepackage{graphicx}
\usepackage{amsmath}
\usepackage{amssymb}

\usepackage{tabularx}
\usepackage{algorithmic}
\usepackage{algorithm}
\usepackage{array}
\usepackage{multirow}
\usepackage{booktabs}
\usepackage[nocompress]{cite}
\usepackage{authblk}
\newcolumntype{C}[1]{>{\centering\arraybackslash}m{#1}}
\newcolumntype{L}[1]{>{\arraybackslash}m{#1}}
\newcolumntype{P}[1]{>{\centering\arraybackslash}p{#1}}
\usepackage{pifont}%
\newcommand{\cmark}{\ding{51}}%
\newcommand{\xmark}{\ding{55}}%

\usepackage[pagebackref=true,breaklinks=true,letterpaper=true,colorlinks,bookmarks=false]{hyperref}

\iccvfinalcopy %

\def\iccvPaperID{2312} %

\ificcvfinal\fi

\begin{document}

\title{MVPSNet: Fast Generalizable Multi-view Photometric Stereo}
\author{
    Dongxu Zhao$^{1}$\hspace{0.5em}
    Daniel Lichy$^{2}$\hspace{0.5em}
    Pierre-Nicolas Perrin$^{1}$\hspace{0.5em}
    Jan-Michael Frahm$^{1}$\hspace{0.5em}
    Soumyadip Sengupta$^{1}$\\
    $^{1}$University of North Carolina at Chapel Hill \quad $^{2}$University of Maryland, College Park \\
    {\tt\small \{dongxuz1, pnperrin, jmf, ronisen\}@cs.unc.edu \qquad dlichy@umd.edu}
}

\maketitle
\ificcvfinal\thispagestyle{empty}\fi

\begin{abstract}
\vspace{-1em}
  We propose a fast and generalizable solution to Multi-view Photometric Stereo (MVPS), called MVPSNet. The key to our approach is a feature extraction network that effectively combines images from the same view captured under multiple lighting conditions to extract geometric features from shading cues for stereo matching. We demonstrate these features, termed  `Light Aggregated Feature Maps' (LAFM), are effective for feature matching even in textureless regions, where traditional multi-view stereo methods fail. Our method produces similar reconstruction results to PS-NeRF, a state-of-the-art MVPS method that optimizes a neural network per-scene, while being 411$\times$ faster (105 seconds vs. 12 hours) in inference. Additionally, we introduce a new synthetic dataset for MVPS, sMVPS, which is shown to be effective to train a generalizable MVPS method.
\vspace{-2em}
\end{abstract}

\vspace{-0.5em}
\section{Introduction}
\vspace{-0.5em}

3D reconstruction of an object can be achieved either through camera viewpoint variations, Multi-view Stereo (MVS), or by lighting direction variations, Photometric Stereo (PS). Both MVS and PS have relative strengths and weaknesses. While MVS succeeds in obtaining accurate global shapes, it suffers in textureless regions due to poor feature matching, often resulting in reconstructions that lack local details. On the other hand, PS produces accurate local details, even in textureless regions, by using shading information but fails to reconstruct accurate global shapes. In this paper, we focus on the problem of Multi-view Photometric Stereo (MVPS) where both camera viewpoint and lighting direction variations are used to accurately reconstruct global and local details of a 3D shape, even in textureless regions.

3D reconstruction techniques that produce high-quality results using only viewpoint variations (MVS) rely on test-time optimization, often by training neural networks per scene \cite{yariv2020multiview,zhang2021nerfactor,srinivasan2021nerv}. These methods are computationally inefficient, typically taking hours of compute time on a high-end GPU for each object. Existing MVS methods \cite{yao2018mvsnet,casmvsnet,sun2021neucon} that focus on computational efficiency employ feed-forward neural networks  that are efficient but fail to produce high-quality details, especially in textureless regions. Existing MVPS approaches can produce high-quality reconstructions but require computationally inefficient per-scene training or optimization \cite{kaya2022neural,kaya2022uncertainty,kaya2023multi,yang2022ps}. Sometimes additional manual efforts and carefully crafted refinement steps are also needed \cite{park2016robust,li2020multi}. In contrast, we propose an efficient feed-forward neural architecture, MVPSNet, that can generalize to unseen objects and achieve similar reconstruction quality to that of per-scene optimization techniques while being computationally efficient during inference.

We design MVPSNet by taking inspiration from various deep MVS architectures \cite{yao2018mvsnet,casmvsnet,ding2022transmvsnet,cao2022mvsformer,giang2021curvature} that are generalizable, computationally efficient, and can operate on high-resolution images. However, these approaches fail in textureless regions, and their reconstructed meshes often lack details. We choose the CasMVSNet \cite{casmvsnet} architecture as our feature matching module, which has been repeatedly used by various MVS pipelines \cite{ding2022transmvsnet,cao2022mvsformer,giang2021curvature} for its simplicity and efficiency, and augment it to effectively incorporate lighting variation cues for better prediction of 3D shapes. To our knowledge, we are the first to propose a feed-forward generalizable approach to Multi-view Photometric Stereo.

We introduce a multi-scale feature representation, called Light Aggregated Feature Maps (LAFM), whose role is to extract detailed geometric features from images by utilizing lighting variations. For brevity, we define \textbf{Multi-light Images} as a collection of images taken from the same viewpoint under different directional lighting conditions. Our intuition is that LAFM can efficiently aggregate shading patterns from multi-light images, by creating an `artificial shading texture' in the textureless region. Multiscale LAFMs will then be used to construct a sequence of Cost Volumes to match features across sparse viewpoints in order to predict a depth map for each viewpoint. We also predict surface normals from LAFM for each viewpoint, enabling us to capture features related to high-frequency local details. We further show that the surface normal predicted by using LAFM can be used in addition to the depth maps to produce a more detailed mesh than using the depth maps alone.

\begin{figure*}[t]
\begin{center}
\includegraphics[width=0.95\linewidth]{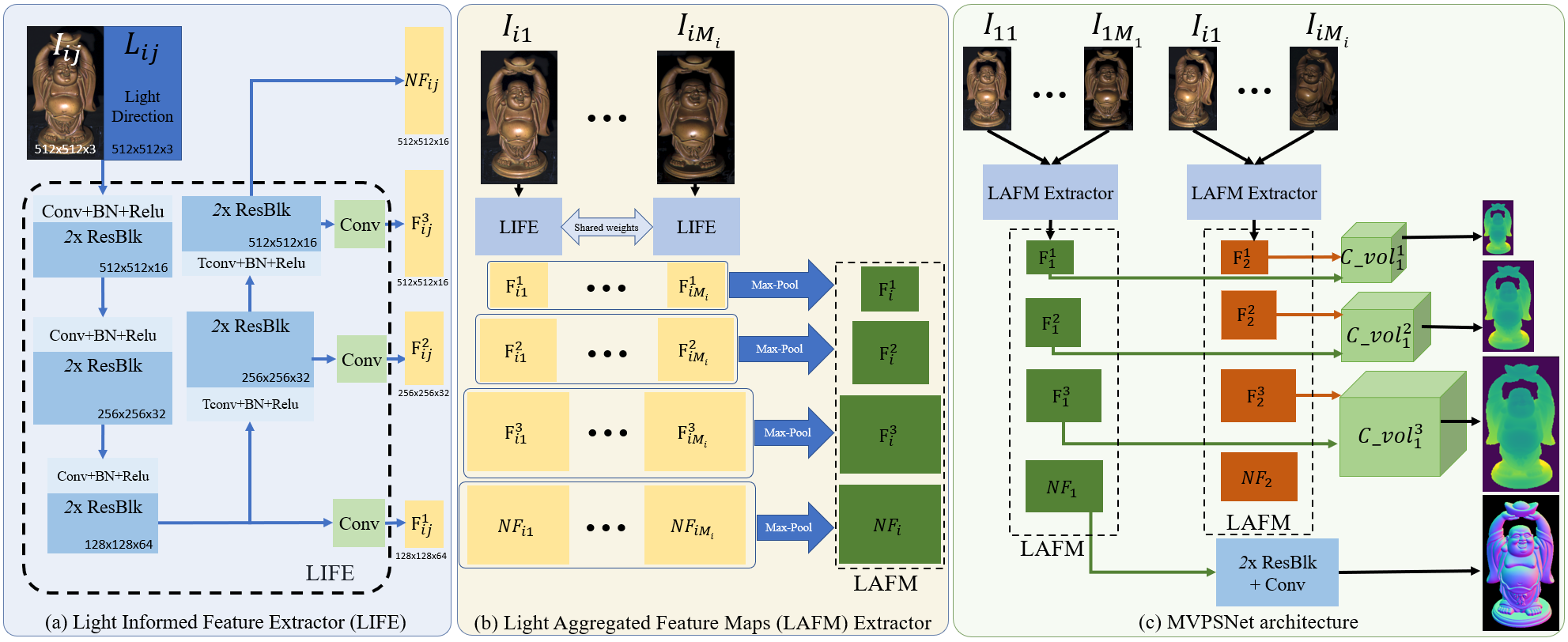}
\end{center}
\vspace{-0.5em}
   \caption{Overview of our network  architecture. (a) Light Informed Feature Extractor (LIFE) produces a multi-scale feature representation; (b) Light Aggregated Feature Maps (LAFM) Extractor aggregates these feature across images of varying lighting conditions but same view; and (c) LAFM is used to create Cost Volume and predict depth map, similar to CasMVSNet \cite{gu2020cascade}, in addition to normal map.}
\label{fig:arthitecture}
\vspace{-1em}
\end{figure*}

To train the proposed MVPSNet architecture, we introduce a new synthetic MVPS dataset. Our synthetic dataset consists of shapes from sculpture dataset \cite{sculpture_data} and random compositions of primitive shapes generated by \cite{xu2018deep}. We render these shapes with spatially varying Cook-Torrance BRDF under different camera viewpoints and lighting directions. We train MVPSNet on these rendered images with ground-truth supervision over predicted depth and surface normal maps. The trained model generalizes to real-world test scenes from DiLiGenT-MV \cite{li2020multi} dataset. We show that simply re-training CasMVSNet on our dataset improves reconstruction quality over the pre-trained model on DiLiGenT-MV by 32\%, proving the effectiveness of our synthetic MVPS dataset for generalization.

We evaluate our approach on the only publicly available MVPS benchmark, the DiLiGenT-MV \cite{li2020multi} dataset. Compared to the state-of-the-art MVPS technique, PS-NeRF \cite{yang2022ps}, which optimizes a neural network per-scene, our proposed MVPSNet is $\sim$411$\times$ faster (105 seconds vs 12 hours) while producing similar reconstruction quality (L1 Chamfer distance of 0.82 vs 0.81, F-score on L2 distance of 0.985 vs 0.983). We further show that adding LAFM features significantly improves reconstruction quality over CasMVSNet by 34\% in L1 Chamfer distance. We also observe that refining the reconstructed mesh derived from depth maps with predicted surface normals from LAFM features improves reconstruction quality as shown in Fig \ref{fig:zoom_qual_study}.

In summary, the key contributions of this paper include: $\bullet$ Light Aggregated Feature Maps (LAFM) that can efficiently utilize multi-light images to extract detailed geometric features, especially in textureless regions. The surface normal predicted from LAFM also improves mesh reconstruction quality. $\bullet$ A synthetic MVPS dataset for training generalizable MVPS methods, which also improves CasMVSNet by 32\%. $\bullet$ A fast and generalizable Multi-view Photometric Stereo pipeline that is 411$\times$ faster while producing similar reconstruction accuracy compared to state-of-the-art per-scene optimization approach \cite{yang2022ps}.

\vspace{-1em}
\section{Related work}
\vspace{-0.5em}

\textbf{Multi-View Stereo (MVS).} MVS is a 3D reconstruction technique that utilizes multiple images captured from different viewpoints. While various techniques for MVS have been proposed, one commonly used approach that is relevant to our work involves constructing cost volumes similar to Plane Sweeping Algorithm \cite{collins1996planesweepingstereo}. To create a cost volume, features are matched across neighboring viewpoints, and the quality of the features plays a critical role in the final reconstruction quality. Traditional methods \cite{kang2001handling, strecha2004wide, strecha2006combined, goesele2007multi, furukawa2009accurate, bleyer2011patchmatch, furukawa2010towards, schonberger2016pixelwise} use human-defined or hand-crafted image processing operators to extract feature maps. With recent advances in deep learning, features are extracted with a deep neural network to build cost volumes and then predict per-view depth maps \cite{ji2017surfacenet, yao2018mvsnet, yao2019recurrent, luo2019p, gu2020cascade, xu2020learning} or disparity map \cite{huang2018deepmvs}.

The most relevant previous works are MVSNet \cite{yao2018mvsnet} and its variations. MVSNet \cite{yao2018mvsnet} uses homography to warp feature maps and a 3D CNN to regularize cost volumes. CasMVSNet \cite{gu2020cascade} outperforms MVSNet in terms of accuracy and efficiency by building the 3D cost volume in a cascaded manner. TransMVSNet \cite{transmvsnet} builds upon CasMVSNet and adopts a transformer to consider intra-image and inter-image feature interactions, which further improves the result of CasMVSNet.

\textbf{Photometric Stereo (PS).} PS (introduced in \cite{woodham1980photometric}) uses lighting variation to reconstruct 3D shapes from a single viewpoint (see \cite{diligent_data} for surveys). Calibrated PS approaches, like Chen \etal \cite{chen2018ps}, train a neural network to predict surface normals using data with known lightings. Uncalibrated PS approaches \cite{chen2019self,chen2020learned,chen2020deep} first predict the lighting parameters before solving for surface normals. While most PS works use a large number of images for inference, some use fewer \cite{Lichy_2021_CVPR, pstransformer}, or even one image \cite{sengupta2018sfsnet, li2018learning, Boss2020-TwoShotShapeAndBrdf, lichy2022fast}  (often called Shape from Shading). PS approaches are mostly based on feed-forward networks that generalize and can produce near real-time inference with low computational cost \cite{Lichy_2022_CVPR}.

\begin{table}
\scriptsize
\begin{center}
\begin{tabular}{l|cc}
\hline
Method & Generalizable  & Mesh Reconstruction\\
\hline
PJ16 \cite{park2016robust} & \xmark & Base mesh+displacement map \\
LZ20 \cite{li2020multi} & \xmark & 3D points+PSR\cite{kazhdan2006poisson}+Optimization\cite{nehab2005efficiently}\\
BKW22\cite{kaya2022neural} & \xmark & MLP+Marching Cube\cite{marching_cube} \\
BKC22\cite{kaya2022uncertainty} & \xmark & MLP+Marching Cube\cite{marching_cube}  \\
PS-NeRF \cite{yang2022ps}  & \xmark & MLP+MISE\cite{mescheder2019occupancy}\\
BKW23\cite{kaya2023multi} & \xmark & MLP+Marching Cube\cite{marching_cube} \\
\hline
Ours & \cmark & 3D Points+Screened Poisson\cite{kazhdan2013screened}\\
\hline
\end{tabular}
\end{center}
\vspace{-0.5em}
\caption{\small{Comparison of our method with prior MVPS methods.}}
\label{table:MVPS_comparison}
\vspace{-2em}
\end{table}

\textbf{Multi-View Photometric Stereo (MVPS).} MVPS was initially proposed in \cite{hernandez2008multiview} by combining PS with object silhouettes to reconstruct textureless shiny objects with fine details. However, this method only works well for specific parametric BRDF models \cite{kaya2022uncertainty}. Later, Li \etal \cite{zhou2013multi, li2020multi} propose to get iso-depth contours from PS images and sparse 3D points using structure-from-motion, which are propagated to recover complete 3D shape. Park \etal \cite{park2013multiview, park2016robust} use a planar mesh parameterization technique to parameterize a coarse mesh from MVS and take advantage of this 2D parameter domain to perform MVPS. However, these traditional MVPS methods require an initial 3D reconstruction and their performance is sensitive to it. Since they consist of multiple steps, careful execution or expert interventions are often performed to get good results \cite{park2016robust,li2020multi}.

Recently, inspired by NeRF \cite{mildenhall2021nerf}, various algorithms have been proposed that optimize a neural network per-scene for MVPS. 
Kaya \etal \cite{kaya2022neural} train a deep PS network first and condition the color rendering in NeRF \cite{pixelnerf} on normals predicted from PS. The reconstructed mesh, however, exhibits multiple artifacts. 
The authors in \cite{kaya2022uncertainty} propose to train a deep PS network and a deep MVS network extended with uncertainty estimation separately, and use these to fit an SDF represented by an MLP. 
To further enable reconstruction on anisotropic and glossy objects, Kaya \etal \cite{kaya2023multi} add a neural volume rendering module to the MLP used in \cite{kaya2022uncertainty} to better fuse PS and MVS measurements. PS-NeRF \cite{yang2022ps} solves the task of jointly estimating the geometry, materials and lights. It first regularizes the gradient of a UNISURF \cite{oechsle2021unisurf} with estimated normals from PS, and then uses separate MLPs to explicitly model surface normal, BRDF, lights, and visibility which are optimized based on a shadow-aware differentiable rendering layer. Recent works have also used physically based differentiable rendering either inside a NeRF framework \cite{asthana2022neural} or separately for optimization \cite{zhang2021physg}. However, these methods optimize models for each object independently, thus they do not generalize and are computationally inefficient. 

In contrast, to our knowledge, we are the first to provide a generalizable solution to Multi-view Photometric Stereo by training on our proposed synthetic MVPS dataset. 

\vspace{-0.5em}
\section{Our approach}
\vspace{-0.5em}

\subsection{Problem Setup}
\vspace{-0.5em}

We focus on the problem of calibrated multi-view photometric stereo, i.e. the locations of the light sources and the cameras are known a priori (calibrated prior to capture). The input data consists of a set of multi-light images of an object captured from multiple views. 

Concretely, for the $i$-th view we have $M_i$ images with varying lighting directions $l_{ij}$, denoted as $I_{ij}$. We  refer to the collection $\{I_{i1},...,I_{iM_i}\}$ as the multi-light images for the i-$th$ view. For each view, we are given camera intrinsic matrix $K_i$ and camera extrinsic parameters in the form of a rotation matrix, $R_i$, and a translation vector, $t_i$. Similar to virtually all MVS methods, we assume that we are provided with the depth range for each view.

\vspace{-0.5em}
\subsection{Motivation}
\vspace{-0.5em}

Our approach follows a long line of work in Multi-view Stereo which uses Plane Sweep to construct a Cost Volume and predicts a depth map aligned with a reference image. Recent advances in Plane Sweep Stereo using deep neural networks, especially CasMVSNet \cite{gu2020cascade}, have proven to be generalizable across scenes and can predict high-resolution reconstruction in a matter of seconds. In contrast to previous Multi-view Photometric Stereo approaches, which optimize a neural network per scene, our goal is to produce a generalizable solution. Thus we aim to build on the Plane Sweep stereo architecture proposed in CasMVSNet \cite{gu2020cascade}.

CasMVSNet learns a deep image feature encoder for extracting representative features that can aid in feature matching across multiple views and create a better Cost Volume. However, these features are often ambiguous for non-textured regions and often fail to preserve the geometric details. We believe incorporating lighting variations along with viewpoint variations can lead to better features, which in turn will produce better Cost Volumes and depth maps.

To this end, we introduce `Light Aggregated Feature Maps' (LAFM), whose goal is to extract detailed geometric features, even for textureless regions, by jointly learning to aggregate feature maps over all images captured from a single viewpoint and multiple lighting directions. To obtain effective features that capture geometric details we use LAFM to regress surface normals. We show that LAFM provide superior information for stereo-matching than single-lighting feature maps (as used in CasMVSNet) and thus provide a better reconstruction. We also show that surface normals predicted using LAFM can be used during mesh reconstruction to improve quality over depth maps alone.

Our approach proceeds in three stages. We first extract multi-scale feature representation from each image, along with its lighting directions, using a shared neural network, `Light Informed Feature Extractor' (LIFE). Then we aggregate features extracted by LIFE across all images captured under the same viewpoint but different lighting conditions using a max-pooling operation to form `Light Aggregated Feature Maps' (LAFM). We can then create a Cost Volume for the reference view by matching LAFM of the reference view with all the LAFM from neighboring views. Finally, for each reference view, we predict a depth map using cost volume regularization and a surface normal map from the LAFM. We train our system in a multi-task learning framework with supervised losses over depth and normal predictions. In the following sections, we provide the details of our MVPSNet pipeline. We provide an overview of MVPSNet architecture in Figure \ref{fig:arthitecture}.

\vspace{-0.5em}
\subsection{Light Aggregated Feature Maps (LAFM)}%
\vspace{-0.5em}

We introduce Light Aggregated Feature Maps (LAFM) that provide geometrically distinct multi-scale features for Cost Volume creation in a Plane Sweep Stereo approach. Our key observation is that the multi-light images provide us with important information for feature matching. For textureless regions, the variation in shading (including cast and attached shadows) created by different lighting directions can be interpreted as `artificial' textures. Thus the role of LAFM for textureless regions is to capture the variation in shading as an `artificial' texture that can be used for feature matching across different viewpoints. We also use LAFM to predict surface normal maps, enabling it to capture geometric details required for producing high-quality normal maps. Hence LAFM can capture better features for textureless regions and for reconstructing details, which were absent in the usual deep image features used in deep multi-view stereo algorithms.

We first define a multi-scale feature extractor, Light Informed Feature Extractor (LIFE), that takes an image $I_{ij}$ associated with its lighting direction $L_{ij}$ as input and produces features maps at three different scales $F^1_{ij},F^2_{ij},F^3_{ij}$ at resolutions $1/4,1/2,1$ of the input resolution, and another feature map $NF_{ij}$ that will be used for normal prediction. The network architecture of LIFE is shown in Fig. \ref{fig:arthitecture}(a) and will be discussed in details in the supplementary material.

\setlength{\abovedisplayskip}{0pt} \setlength{\abovedisplayshortskip}{0pt}
\setlength{\belowdisplayskip}{0pt} \setlength{\belowdisplayshortskip}{0pt}
\vspace{-1em}
\begin{equation}
    NF_{ij}, F^1_{ij},F^2_{ij},F^3_{ij} = LIFE(I_{ij},l_{ij};\theta)
    \label{eq:life}
\end{equation}
Note $L_{ij}$ is of the same resolution as $I_{ij}$ by simply repeating the same 3-dimensional lighting vector at each pixel. 

Then we extract these multi-scale features for every image captured under the same viewpoint and different lighting conditions, $\{I_{i1}, ... , I_{iM_i}\}$, using the same shared encoder LIFE. Let the feature maps obtained from these images be denoted as: $\{NF_{ij}, F^1_{ij},F^2_{ij},F^3_{ij}\}$, $j=1, ..., M_i$. We create `Light Aggregated Feature Maps' (LAFM) from these multi-scale feature representations by simply performing a max-pooling operation for each scale, as proposed in \cite{chen2019self}, as:
\begin{align}
     F_i^s &= \max_j F^s_{ij}, \quad \forall s=1,2,3\\
    NF_i &= \max_j NF_{ij}.
\end{align}

Thus for multi-light images we obtain LAFM as $LF_i = \{NF_i,F^1_{i},F^2_{i},F^3_{i}\}$.

The features at 3 scales $F^1_{i},F^2_{i},F^3_{i}$ are then used to build cost volumes using differentiable homography warping, which we will talk about in detail in Section 4.1. The normal feature $NF_i$ is fed into a lightweight normal regression network to predict per-view normal map, as shown in Figure \ref{fig:arthitecture}(c). With the supervision from normal information and depth information, our LAFM benefit from the advantages of both MVS and PS which are good at global shape modeling and high-frequency component reconstruction, respectively.

\vspace{-0.5em}
\subsection{Cost Volume and Depth Map Prediction}
\vspace{-0.5em}

Given Light Aggregated Feature Maps (LAFM), $LF_{i}$, for each view $i$, we aim to build a cost volume for each reference view by selecting a set of source views with sufficient overlap. We adopt the multi-scale cost volume construction proposed in CasMVSNet \cite{gu2020cascade}, where the plane sweep is first performed at a low resolution and then at higher resolutions. Depth estimated from the previous step is used for generating depth proposals for the next step. Multi-scale cost volume reconstruction and depth map prediction follow the following steps.

\textbf{Step 1: Depth hypothesis generation.} We generate hypothesis depths for each pixel based on the lower resolution depth estimated at the previous resolution. We store these in $h$, where
\vspace{-0.5em}
\begin{equation}
    h(u,v,w) = Up(D^{s-1})(u,v) + \Delta_s (\frac{w}{N_s-1} - \frac{1}{2}).
\end{equation}
Here $h(u,v,w)$ is the $w$-th depth hypothesis at pixel $(u,v)$. $Up(D^{s-1})$ is the depth map at the previous lower resolution upsampled to the current resolution. $\Delta_s$ is the length of the depth interval we are searching at scale $s$. $N_s$ is the number of hypothesis depths at the current scale.

\textbf{Step 2: Cost-Volume construction.} The cost volume is a way of robustly searching for matches between a point $u,v$ on a reference image $I_r$ and a point on the corresponding epipolar line in the source images $I_{s_k}$. Concretely, consider a pixel $(u,v)$ in the reference image. For every hypothesis depth $d$, we get a corresponding point in the source image on the epipolar line for $(u,v)$. We denote this point by $(u',v') = \text{warp}_{rs_k}(u,v,d)$ where

\vspace{-1em}
\begin{equation}
\small{
    \left[\begin{array}{c}
         u'  \\
         v'  \\
         1
    \end{array} \right]
    \sim
    K_{s_k}R_{s_k}^T\left[ \left( R_rK_r^{-1}d \left[\begin{array}{c}
         u  \\
         v  \\
         1
    \end{array} \right] + t_r\right) - t_s \right].}
\end{equation}

We then construct a per-image volume:
\begin{equation}
    \text{F\_{vol}}_i^s(u,v,w) = F_i^s(warp(u,v,h(u,v,w))),
\end{equation}
where $i$ runs over the reference and source views i.e. $i \in \{r,s1,...,s_k\}$. These volumes are then aggregated into a single cost volume by taking their variance, which checks for the photo-consistency of the depth proposal $d$ for pixel $(u,v)$ in the reference image and the corresponding pixels in the warped sources images $s_k$:
\begin{equation}
    \text{agg\_vol}^s(u,v,w) = \text{var}_i (\text{F\_{vol}(u,v,w)}_i^s).
\end{equation}

\textbf{Step 3: Cost-Regularization.} In this step we pass the aggregated volume through a 3D convolutional network and take a softmax to convert it to a match probability, using
\begin{equation}
    \text{prob\_vol} = \text{soft\_max}_w (\text{reg\_net}^s(\text{agg\_vol}^s)).
\end{equation}

\textbf{Step 4: Regression.} We take the expectation of the hypothesis depths over the match probability given by the probability volume to obtain the depth at the current scale: 
\begin{equation}
     D^s(u,v) = \sum_w \text{prob\_vol}(u,v,w)h(u,v,w).
\end{equation}

This whole process is summarized in algorithm \ref{algo:casMVSAlgorithm}

\vspace{-1em}
\begin{algorithm}[H]
\caption{MVPSNet Algorithm}
\label{algo:casMVSAlgorithm}
\begin{algorithmic}[1]
\STATE $NF_{ij}, F^1_{ij},F^2_{ij},F^3_{ij} = \text{LIFE}(I_{ij},l_{ij};\theta)$
\STATE $NF_i = \max_j NF_{ij}$; $F^s_i = \max_j F^s_{ij} \quad \forall s=1,2,3.$
\STATE $N_i = \text{normal\_regression\_net}(NF_i)$

\STATE $D^0(u,v) = (\text{max\_depth}+\text{min\_depth})/2$
\FOR {s = 1 to 3} 
    \STATE $h(u,v,w) = Up(D^{s-1}(u,v)) + \Delta_s (\frac{w}{N_w-1} - \frac{1}{2})$
    \STATE $\text{F\_{vol}}_i^s(u,v,w) = F_i^s(warp(u,v,h(u,v,w)))$
    \STATE $\text{agg\_vol}^s(u,v,w) = \text{var}_i (\text{F\_{vol}}_i^s)$
    \STATE \text{prob\_vol} = $\text{soft\_max}_w (\text{reg\_net}^s(\text{agg\_vol}^s))$
    \STATE $D^s(u,v) = \sum_w \text{prob\_vol}(u,v,w)h(u,v,w)$
\ENDFOR
\end{algorithmic}
\end{algorithm}
\vspace{-1em}
Once we have depth and surface normal for each view, our mesh reconstruction pipeline consists of three steps: depth filtering, lifting depth and normal maps to point cloud, and reconstructing mesh using Screened Poisson \cite{kazhdan2013screened} (See supplementary materials for details).

\vspace{-0.5em}
\subsection{Synthetic MVPS dataset}
\vspace{-0.5em}

A key component of our method is that we can learn better features for stereo matching, especially in textureless regions, by learning features that incorporate multi-lighting cues. However, there is no existing MVPS dataset that is large enough for neural network training. Therefore, we generate a large-scale synthetic dataset, sMVPS, consisting of two sub-datasets sMVPS-sculpture (800 train scenes/4 test scenes) and sMVPS-random (1000 train scenes/20 test scenes).

sMVPS-sculpture consists of objects from the sculpture dataset \cite{sculpture_data} while sMVPS-random includes objects composed of random  primitives from \cite{xu2018deep}. The objects were generated following the method of \cite{lichy2021shape} with spatially varying Cook-Torrance BRDF.
We render images from 20 viewpoints surrounding the object approximately every 18$^\circ$ plus random jitter in position. For each view we render 10 randomly chosen directional light sources sampled uniformly on a 45$^\circ$ spherical cap centered at the camera's optical axis. In addition to the images, we render ground truth normals, depth, albedo, and roughness.

\begin{figure}[t]
\begin{center}
   \includegraphics[width=0.98\linewidth]{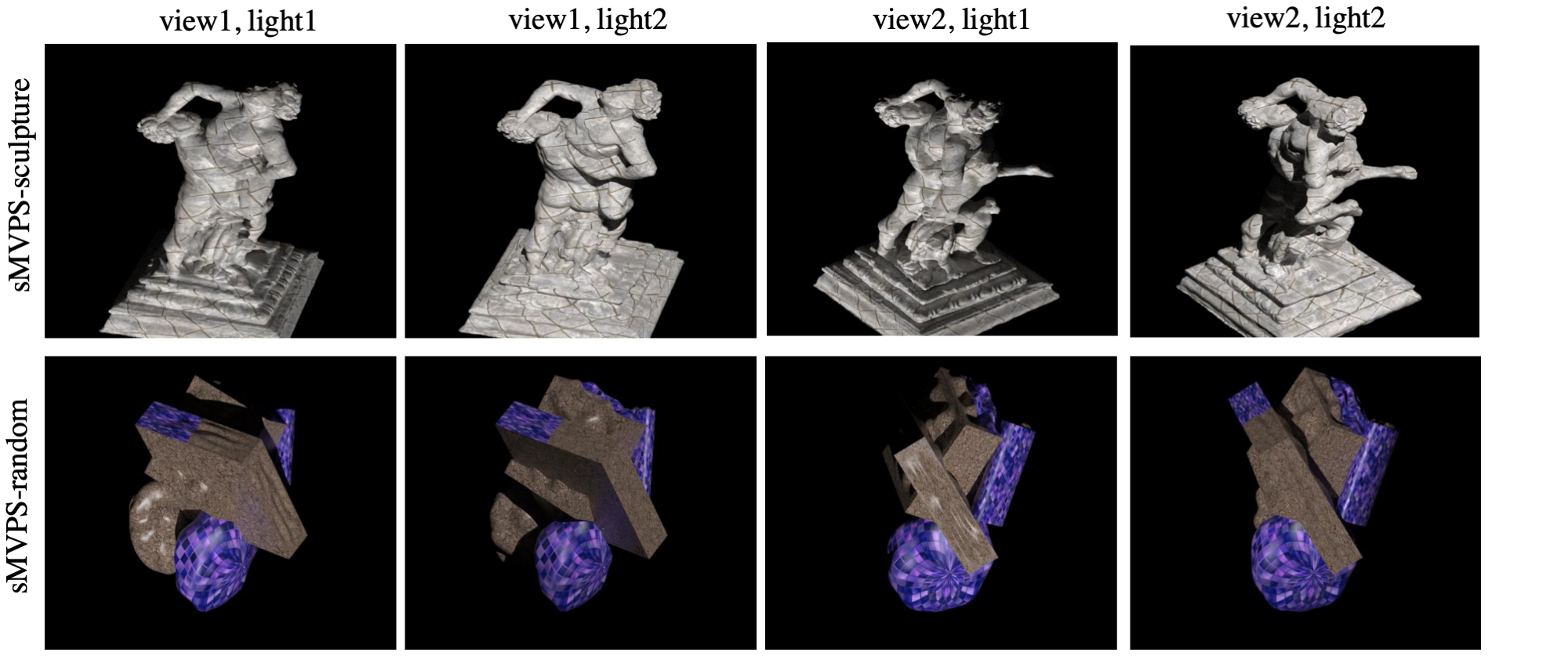}
\end{center}
\vspace{-1em}
   \caption{Example images from proposed synthetic MVPS dataset.}
\label{fig:synthetic_data}
\label{fig:onecol}
\vspace{-1.5em}
\end{figure}

\vspace{-0.5em}
\subsection{Training MVPSNet}
\vspace{-0.5em}

     We train MVPSNet with supervised loss over surface normal and depth using the ground-truth created with synthetic sMVPS dataset. For each reference view, we use 2 source views. And we randomly choose 3 lights out of 96 to train our model. The total loss is defined as:
     \begin{align}
     L_{mvps} &= \lambda_{d} \cdot Loss_{d} + \lambda_{n} \cdot Loss_{n},   \\ 
     L_{d} &= \sum_{s=1}^{3} \lambda_{ds} \cdot L_{ds}, \forall s=1,2,3
     \end{align}
    where $L_{ds}$ and $L_{n}$ refer to the depth loss for scale $s$ and normal loss, respectively. For loss weights, we set $L_{n}=1$ and $L_{d}=10$. The weights of each scale, $\lambda_{ds}=1$, for $s=1,2,3$.

\begin{table*}
\footnotesize
\begin{center}
\begin{tabular}{l|P{3em}P{3em}|P{3em}P{3em}P{4em}|P{7em}|P{5em}P{6em}|P{4em}}
\toprule
 &  \multicolumn{5}{c|}{{\small \textbf{Per-scene optimization}}} & \multicolumn{4}{c}{{\small \textbf{Generalizable}}}\\ 
\hline
Category  &  \multicolumn{2}{c|}{Manual Effort} & \multicolumn{3}{c|}{Standalone} & Single-view PS & \multicolumn{2}{c|}{MVS} & MVPS\\ 
\hline
Method  & PJ16\cite{park2016robust} & LZ20\cite{li2020multi} & BKW22\cite{kaya2022neural} & BKC22\cite{kaya2022uncertainty} & PS-NeRF\cite{yang2022ps} & PS-Transformer\cite{pstransformer} &  CasMVSNet\cite{gu2020cascade}- RT & TransMVSNet\cite{transmvsnet}- RT & Ours\\
\hline 
BEAR & 2.54 & 0.73 & 1.01 & 1.01 & \textbf{0.76} & 3.17 &  1.47 & 1.48 & \underline{0.80}\\
BUDDHA & 1.12 & 0.97 & 2.68 & 1.15 & \textbf{0.86} & 4.09 &  1.26 & 1.10 & \underline{1.07}\\
COW & 1.14 & 0.39 & 1.09 & \underline{0.76} & \textbf{0.75} & 3.04  & 1.27 & 1.05 & 0.77\\
POT2 & 3.21 & 0.67 & 1.54 & 1.40 & \textbf{0.76} & 3.05  & 1.46 & 1.05 & \underline{0.82}\\
READING & 1.30 & 0.66 & 1.97 & 0.84 & 0.92 & 3.60  & \underline{0.75} & 0.76 & \textbf{0.66} \\
\hline
AVERAGE & 1.86 & 0.69 & 1.66 & 1.03 & \textbf{0.81} & 3.39  & 1.24 & 1.09 & \underline{0.82}\\
\hline
Recon. Time/object & - & - & 7 hrs & ? & 12 hrs & ?  & \textbf{22s} & \underline{52s} & 105s\\
\bottomrule
\end{tabular}
\end{center}
\vspace{-0.5em}
\caption{L1 Chamfer Distance (lower is better) between reconstructed mesh and GT after ICP. `-RT' denotes trained on our synthetic MVPS dataset. For non-manual methods, the best result is shown in bold, 2nd best as underline. LZ20 \& PJ16 involve carefully crafted steps, manual efforts in finding correspondence, and an initial mesh or point cloud.}
\vspace{-1em}
\label{table:3d_results_Chamfer_dist}
\end{table*}

\begin{table*}
\footnotesize
\begin{center}
\begin{tabular}{P{4em}|P{2em}P{2em}|P{3em}P{3em}P{3em}P{4em}|P{7em}|P{5em}P{6em}|P{7em}}
\toprule
 &  \multicolumn{6}{c|}{{\small \textbf{Per-scene optimization}}} & \multicolumn{4}{c}{{\small \textbf{Generalizable}}}\\ 
\hline
Category  &  \multicolumn{2}{c|}{Manual Effort} & \multicolumn{4}{c|}{Standalone} & Single-view PS & \multicolumn{2}{c|}{MVS} & MVPS\\ 
\hline
Method  & PJ16\cite{park2016robust} & LZ20\cite{li2020multi} & BKW22\cite{kaya2022neural} & BKC22\cite{kaya2022uncertainty} & BKW23*\cite{kaya2023multi}& PS-NeRF\cite{yang2022ps} & PS-Transformer\cite{pstransformer} & CasMVSNet\cite{gu2020cascade}-RT & TransMVSNet\cite{transmvsnet}-RT & Ours\\
\hline 
BEAR & 0.551 & 0.986 & 0.928 & 0.934 & 0.965 & \textbf{0.995} & 0.078 & 0.911 & 0.882 & \underline{0.991}\\
BUDDHA & 0.940 & 0.936 & 0.687 & 0.926 & \textbf{0.993} & \underline{0.983} & 0.066  & 0.919 & 0.963 & 0.958\\
COW & 0.918 & 0.990 & 0.937 & 0.986 & \underline{0.987} & 0.986 & 0.140  & 0.914 & 0.941 & \textbf{0.993}\\
POT2 & 0.484 & 0.985 & 0.909 & 0.889 & \underline{0.991} & \underline{0.991} & 0.101  & 0.901 & 0.964 & \textbf{0.994}\\
READING & 0.905 & 0.975 & 0.810 & 0.971 & 0.975 & 0.961 & 0.961  & \underline{0.980} & 0.978 & \textbf{0.988}\\
\hline
AVERAGE & 0.760 & 0.974 & 0.854 & 0.941 & 0.982 & \underline{0.983} & 0.269 & 0.925 & 0.946 & \textbf{0.985}\\
\hline
Recon. Time/object & - & - & 7 hrs & ? & ? & 12 hrs & ?  & \textbf{22s} & \underline{52s} & 105s\\
\bottomrule
\end{tabular}
\end{center}
\vspace{-0.5em}
\caption{F-score with L2 distance and 1mm threshold distance (higher is better) between reconstructed mesh and GT after ICP. `-RT' denotes trained on our synthetic MVPS dataset. For non-manual methods, the best result is shown in bold, 2nd best as underline. LZ20 \& PJ16 involve carefully crafted steps, manual efforts in finding correspondence, and an initial mesh or point cloud. BKW23* code not available, results from the paper.}
\label{table:3d_results_f_score}
\vspace{-2em}
\end{table*}

\vspace{-1em}
\section{Experiments}
\vspace{-0.5em}

\noindent  \textbf{Dataset.} We evaluate our method and conduct ablation study on DiLiGenT-MV \cite{li2020multi} dataset, which is the only benchmark dataset for MVPS tasks and widely used by all previous approaches. It contains images of 5 objects with diverse materials captured from 20 views. 
 For each view, the object is illuminated by one of 96 calibrated point light sources at one time, which gives us 96 images with varying lighting conditions.

\noindent  \textbf{Evaluation Metrics.} We evaluate the quality of recovered meshes using L1 Chamfer distance from PyTorch3D \cite{ravi2020pytorch3d, lassner2020pulsar} and F-score \cite{knapitsch2017tanks} with L2 distance and 1mm threshold distance. The distances in both metrics are computed between the vertices of two meshes. 
The unit is \textit{mm}.

\begin{figure*}[t]
\begin{center}
   \includegraphics[width=0.95\linewidth]{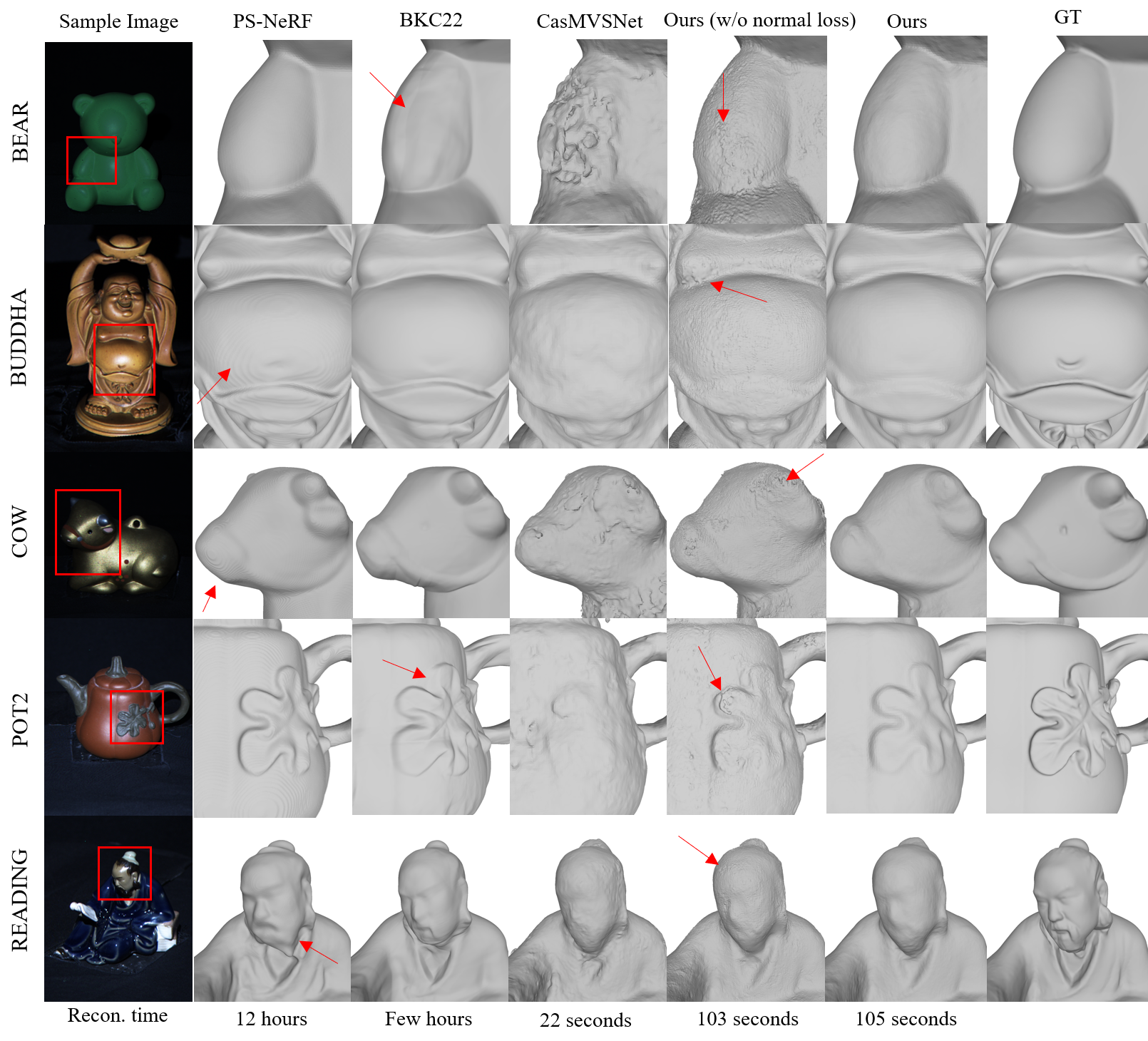}
\end{center}
\vspace{-1em}
   \caption{\small{Qualitative comparison of our method with existing approaches (red arrow highlights artifacts) and ablation studies.}}
\label{fig:zoom_qual_study}
\vspace{-1.5em}

\end{figure*}

\noindent  \textbf{Evaluation details.} Ground truth meshes and meshes of PJ16 \cite{park2016robust} and LZ20\cite{li2020multi} are included in the DiLiGenT-MV dataset \cite{li2020multi}. 
We thank the authors of BKW22 \cite{kaya2022neural} and BKC22 \cite{kaya2022uncertainty} for providing us with their reconstructed meshes. 
For PS-NeRF \cite{yang2022ps} we extract meshes for each object using their released code and unscale it to the scale of the ground truth as suggested in their code. 
The code or reconstructed meshes of BKW23\cite{kaya2023multi} was not released, so we only include their reported F-score on L2 distance in Table \ref{table:3d_results_f_score}. 
To compare with PS-Transformer \cite{pstransformer}, we get normal maps from their pretrained model and integrate normals into depth maps, followed by a depth fusion step after rescaling all depth maps to the ground truth depth scales. 
To compare with CasMVSNet \cite{gu2020cascade} and TransMVSNet \cite{transmvsnet}, we consider both the pretrained models on DTU dataset \cite{dtu} and models retrained on our synthetic dataset, dubbed as CasMVSNet-RT and TransMVSNet-RT. 
Images from 5 views are used to generate each single-view depth map and all 20 depth maps are fused together. For ours, we take images from 5 views and 10 lightings conditions for each view, along with corresponding light directions, as input to generate single-view depth maps and all 20 depth maps are fused for full recovery.
Since there is no image showing the bottoms of objects in the dataset, following BKW22 \cite{kaya2022neural} and BKC22 \cite{kaya2022uncertainty}, we remove points that are located lower than +5 on the z-axis from all reconstructed meshes and the ground truth. 
Similar to most previous approaches \cite{li2020multi, park2016robust, kaya2022neural, kaya2022uncertainty} we also perform a rigid registration using Iterative Closest Point (ICP) \cite{4767965, besl1992method, chen1992object, zhang1994iterative} between the ground-truth and each reconstructed mesh for a fair comparison.

\begin{table}
\footnotesize
\begin{center}
\begin{tabular}{l|P{4em}P{5em}P{3em}P{4em}P{4em}}
\toprule
Method & CasMVSNet & CasMVSNet-RT  & Ours (train 1 light/view) & Ours (train 3 light/view)\\
\hline
BEAR & 2.00 & 1.47 & 1.31 & 0.80 \\
BUDDHA & 1.44 & 1.26 & 1.26 & 1.07 \\
COW & 2.73 & 1.27 &  1.06 & 0.77 \\
POT2 & 1.89 & 1.46 & 1.07 & 0.82 \\
READING & 1.07 & 0.75 & 0.77 & 0.66 \\
\hline
AVERAGE & 1.83 & 1.24  & 1.09 & 0.82\\
\bottomrule
\end{tabular}
\end{center}
\vspace{-1em}
\caption{Ablation: Results are improved by retraining CasMVSNet \cite{casmvsnet} on proposed synthetic dataset (sMVPS). LAFM help in aggregating features across lighting variations, Ours (train 3 light/view) vs Ours (train 1 light/view), and is more accurate than CasMVSNet features, Ours (train 1 light/view) vs  CasMVSNet-RT.}
\vspace{-2em}
\label{table:ablation}
\end{table}

\vspace{-0.5em}
\subsection{Comparison with Existing Approaches.}
\vspace{-0.5em}

We compare our algorithm with approaches that require per-scene optimization or training and with feed-forward generalizable methods. The quantitative result is showed in Table \ref{table:3d_results_Chamfer_dist} and \ref{table:3d_results_f_score}. We also show visual comparison of meshes from representative methods in Figure \ref{fig:zoom_qual_study}.

(i) \textbf{Per-scene Optimization.} Per-Scene optimization methods can also be categorized into:\\
\noindent \textbf{(a) Manual Efforts Needed.} We compare with two traditional multi-stage MVPS methods, PJ16 \cite{park2016robust} and LZ20 \cite{li2020multi}, which require manual efforts. 
We outperform PJ16\cite{park2016robust} with a clear margin. Although LZ20 \cite{li2020multi} achieves better results than ours, note that both methods, PJ16 \& LZ20, consist of multiple steps with carefully crafted geometric modeling. 
Besides, they require an initial mesh \cite{park2016robust} or point cloud \cite{li2020multi} to build upon and their performance is sensitive to the initialization quality. 
When the initialization step fails in large textureless regions, LZ20 \cite{li2020multi} incorporates manual labeling to establish correspondence across views. In contrast, our pipeline is completely automatic without any manual efforts, does not involve any carefully crafted multi-stage approach, and does not require separate hyper-parameters for individual object.

\noindent \textbf{(b) Standalone Methods.} Recent deep learning-based MVPS methods, including BKW22\cite{kaya2022uncertainty}, BKC22\cite{kaya2022neural}, BKW23\cite{kaya2023multi} and PS-NeRF\cite{yang2022ps}, are simpler and easier to adopt. However, like traditional methods \cite{park2016robust, li2020multi}, they still optimize one model for each object individually, resulting in low computational efficiency. 
In contrast, although our model is trained only on synthetic data, we outperform some per-scene optimized methods \cite{kaya2022neural, kaya2022uncertainty} and get comparable results as state-of-the-art, PS-NeRF \cite{yang2022ps}. Furthermore, even though PS-NeRF\cite{yang2022ps} recovers high-quality meshes with details, its recovered surfaces contain iso-contour pattern artifacts, \eg see the red arrows in Fig. \ref{fig:zoom_qual_study} on BUDDHA and COW, and often incorrect shapes, \eg READING. Note that, we could not report L1 Chamfer distance on BKW23\cite{kaya2023multi}, since the code is unavailable, but we show in Table \ref{table:3d_results_f_score} that our method is slightly better than BKW23 in F-score.

(ii) \textbf{Generalizable.} We compare our method with two categories of generalizable methods:\\
\noindent \textbf{(a) Single-view Photometric Stereo.} PS-Transformer \cite{pstransformer} is a state-of-the-art PS network, which takes multiple images with the same viewpoint but different lighting conditions as input and generates single-view normal map prediction. To get 3D reconstruction, we integrate each normal map into a depth map and fuse them together. Since the integrated depths are of arbitrary scale, we rescale them to the range of ground truth depth. Inherently, PS methods struggle with global shape modeling and it is challenging to stitch multi-view integrated depth of arbitrary scale, so PS-Transformer \cite{pstransformer} doesn't perform well on full-view recovery. 

\noindent \textbf{(b) Multi-view Stereo.} We also compare our method with CasMVSNet \cite{gu2020cascade} and TransMVSNet \cite{transmvsnet}, which takes a single lighting image for each view. For fairness, we retrain both methods using our synthetic dataset with suggested hyper-parameters in original papers . We observe that lighting information can largely improve both accuracy and quality, quantitatively and qualitatively.
On textureless objects, \eg BEAR, and textureless regions, \eg belly of BUDDHA, MVS alone gets noisy and rough surfaces. Moreover, our meshes have more high-frequency details that MVS alone may struggle with, \eg the texture on POT2. This is because our LAFMs are supervised with normal maps so they can learn the high-frequency components.

\vspace{-0.5em}
\subsection{Computational efficiency}
\vspace{-0.5em}

While our method outperforms some per-scene optimized methods \cite{kaya2022neural, kaya2022uncertainty} and produces comparable results to state-of-the-arts \cite{yang2022ps,kaya2023multi},  the key advantage of our method is that it is fast, generalizable, and computationally efficient. Thus we analyze to the best of our abilities the inference time of various MVPS algorithms compared in this paper.

$\bullet$ LZ20\cite{li2020multi}: This algorithm takes \textit{117 minutes} per object, without considering the time required for initializing a point cloud or any manual efforts. $\bullet$ BKW22\cite{kaya2022neural}: takes 7 hours to train per object. $\bullet$ BKC22\cite{kaya2022uncertainty}, BKW23\cite{kaya2023multi}: Since the authors did not mention the time required for training these algorithms it is not possible to provide an exact estimate. However, these approaches are based on MLPs, which take hours to train. $\bullet$ PS-NeRF \cite{yang2022ps}: takes \textit{12 hours} to train per object. $\bullet$ CasMVSNet\cite{casmvsnet}:, in contrast, takes only 22 seconds per object, including obtaining depth maps for each view using 5 views (1 reference view and 4 source views) and 1 lighting per view (5 images processed for estimating a depth map), fusing depth maps from all 20 views to a point cloud, computing normals for vertices in the point cloud and adopting Screened Poisson \cite{kazhdan2013screened} to recover a mesh from the point cloud. $\bullet$ MVPSNet (Ours): takes a total of \textit{105 seconds} to create a mesh, including steps of obtaining depth maps for each view using 5 views (1 reference view and 4 source views) and 10 lightings per view (50 images processed in total), fusing depth maps from all 20 views to a point cloud, and adopting Screened Poisson \cite{kazhdan2013screened} to recover a mesh from the point cloud. 

In summary, we are around 240$\times$ faster than BKW22 \cite{kaya2022neural} and around 411.4$\times$ faster than PS-NeRF \cite{yang2022ps} ignoring their mesh extraction time.

\vspace{-0.5em}
\subsection{Ablation study}
\vspace{-0.5em}

The key contribution of this work includes: (a) our synthetic MVPS dataset, sMVPS-sculpture and sMVPS-random, and (b) `Light Augmented Feature Maps' (LAFM) for more accurate mesh reconstruction. Here we design experiments to analyze impacts of these contributions. %

\textbf{Synthetic Data (sMVPS-sculpture and sMVPS-random).} To illustrate the effectiveness of our synthetic data, we evaluate the performance of a CasMVSNet\cite{gu2020cascade} model trained on DTU dataset \cite{dtu}, and compare it with the CasMVSNet-RT model trained on our sMVPS dataset. 
The L1 Chamfer distance metric is reported in Table \ref{table:ablation}. We observe that training on our synthetic dataset improves reconstruction quality by 32.2\%, proving the effectiveness of our proposed data for MVPS reconstructions.
See supplementary materials for comparison between pretrained TransMVSNet \cite{transmvsnet} and TransMVSNet-RT trained on our synthetic dataset.

\textbf{Light Augmented Feature Maps (LAFM).} LAFM plays two key roles in our approach: (i) it aggregates features from images captured with multiple lighting conditions but the same viewpoints, and (ii) it is trained with surface normal loss which helps to preserve high-frequency details in the features. The predicted surface normals are used to further refine the reconstructed mesh.

For understanding the impact of (i), we train our proposed MVPSNet with just a single lighting image per view instead of 3 lightings. Thus LAFMs are only aggregated across 1 image per view. In Table \ref{table:ablation} we observe that using a single lighting per view (`Ours (train 1 light/view)') produces worse results (1.09 vs 0.82) than using 3 lightings per view (`Ours (train 3 light/view)'). However, even a single lighting per view produces better performance than CasMVSNet-RT (1.09 vs 1.24), which is also trained on single lighting per view. This shows that LAFM is effective in both extracting accurate information from just a single image and aggregating shading information across multiple images with varying illumination.

For understanding the impact of (ii), we train our proposed MVPSNet without any surface normal loss, `Ours (w/o normal loss)'. We observe `Ours (w/o normal loss)' is quantitatively comparable to `Ours (w/normal loss)', 0.79 vs 0.82 in L1 chamfer distance and 0.985 vs 0.985 in F-score. However, in Fig. \ref{fig:zoom_qual_study} we observe that the meshes produced by `Ours (w/o normal loss)' are significantly noisier as shown with red arrows.

\vspace{-0.5em}
\section{Conclusion}
\vspace{-0.5em}
In this work, we propose a fast and generalizable approach for MVPS. We introduce Light Aggregated Feature Maps that leverage shading cues from images with the same view under multiple lighting conditions to produce richer features in textureless regions. 
Being trained with normal estimation, LAFM also enable higher quality reconstruction than traditional MVS methods with only little compromise to speed. 
When trained on the synthetic sMVPS dataset we propose, our method produces results comparable to STOA method that is about 400x slower at inference time.

{\small
\bibliographystyle{ieee_fullname}
\bibliography{egbib}

\begin{thebibliography}{10}\itemsep=-1pt

\bibitem{3dtextures}
3d textures.
\newblock \url{https://3dtextures.me/}.
\newblock Accessed: 2020.

\bibitem{dtu}
Henrik Aan{\ae}s, Rasmus~Ramsb{\o}l Jensen, George Vogiatzis, Engin Tola, and
  Anders~Bjorholm Dahl.
\newblock Large-scale data for multiple-view stereopsis.
\newblock {\em International Journal of Computer Vision}, 120:153--168, 2016.

\bibitem{4767965}
K.~S. Arun, T.~S. Huang, and S.~D. Blostein.
\newblock Least-squares fitting of two 3-d point sets.
\newblock {\em IEEE Transactions on Pattern Analysis and Machine Intelligence},
  PAMI-9(5):698--700, 1987.

\bibitem{asthana2022neural}
Meghna Asthana, William~AP Smith, and Patrik Huber.
\newblock Neural apparent brdf fields for multiview photometric stereo.
\newblock {\em arXiv preprint arXiv:2207.06793}, 2022.

\bibitem{besl1992method}
Paul~J Besl and Neil~D McKay.
\newblock Method for registration of 3-d shapes.
\newblock In {\em Sensor fusion IV: control paradigms and data structures},
  volume 1611, pages 586--606. Spie, 1992.

\bibitem{bleyer2011patchmatch}
Michael Bleyer, Christoph Rhemann, and Carsten Rother.
\newblock Patchmatch stereo-stereo matching with slanted support windows.
\newblock In {\em Bmvc}, volume~11, pages 1--11, 2011.

\bibitem{Boss2020-TwoShotShapeAndBrdf}
Mark Boss, Varun Jampani, Kihwan Kim, Hendrik~P.A. Lensch, and Jan Kautz.
\newblock Two-shot spatially-varying brdf and shape estimation.
\newblock In {\em IEEE Conference on Computer Vision and Pattern Recognition
  (CVPR)}, 2020.

\bibitem{cao2022mvsformer}
Chenjie Cao, Xinlin Ren, and Yanwei Fu.
\newblock Mvsformer: Learning robust image representations via transformers and
  temperature-based depth for multi-view stereo.
\newblock {\em arXiv preprint arXiv:2208.02541}, 2022.

\bibitem{chen2019self}
Guanying Chen, Kai Han, Boxin Shi, Yasuyuki Matsushita, and Kwan-Yee~K Wong.
\newblock Self-calibrating deep photometric stereo networks.
\newblock In {\em Proceedings of the IEEE Conference on Computer Vision and
  Pattern Recognition}, pages 8739--8747, 2019.

\bibitem{chen2020deep}
Guanying Chen, Kai Han, Boxin Shi, Yasuyuki Matsushita, and Kwan-Yee~Kenneth
  Wong.
\newblock Deep photometric stereo for non-lambertian surfaces.
\newblock {\em IEEE Transactions on Pattern Analysis and Machine Intelligence},
  2020.

\bibitem{chen2018ps}
Guanying Chen, Kai Han, and Kwan-Yee~K Wong.
\newblock Ps-fcn: A flexible learning framework for photometric stereo.
\newblock In {\em Proceedings of the European Conference on Computer Vision
  (ECCV)}, pages 3--18, 2018.

\bibitem{chen2020learned}
Guanying Chen, Michael Waechter, Boxin Shi, Kwan-Yee~K Wong, and Yasuyuki
  Matsushita.
\newblock What is learned in deep uncalibrated photometric stereo?
\newblock In {\em European Conference on Computer Vision}, 2020.

\bibitem{chen1992object}
Yang Chen and G{\'e}rard Medioni.
\newblock Object modelling by registration of multiple range images.
\newblock {\em Image and vision computing}, 10(3):145--155, 1992.

\bibitem{collins1996planesweepingstereo}
Robert~T Collins.
\newblock A space-sweep approach to true multi-image matching.
\newblock In {\em Proceedings CVPR IEEE Computer Society Conference on Computer
  Vision and Pattern Recognition}, pages 358--363. Ieee, 1996.

\bibitem{ding2022transmvsnet}
Yikang Ding, Wentao Yuan, Qingtian Zhu, Haotian Zhang, Xiangyue Liu, Yuanjiang
  Wang, and Xiao Liu.
\newblock Transmvsnet: Global context-aware multi-view stereo network with
  transformers.
\newblock In {\em Proceedings of the IEEE/CVF Conference on Computer Vision and
  Pattern Recognition}, pages 8585--8594, 2022.

\bibitem{transmvsnet}
Yikang Ding, Wentao Yuan, Qingtian Zhu, Haotian Zhang, Xiangyue Liu, Yuanjiang
  Wang, and Xiao Liu.
\newblock Transmvsnet: Global context-aware multi-view stereo network with
  transformers.
\newblock In {\em Proceedings of the IEEE/CVF Conference on Computer Vision and
  Pattern Recognition}, pages 8585--8594, 2022.

\bibitem{furukawa2010towards}
Yasutaka Furukawa, Brian Curless, Steven~M Seitz, and Richard Szeliski.
\newblock Towards internet-scale multi-view stereo.
\newblock In {\em 2010 IEEE computer society conference on computer vision and
  pattern recognition}, pages 1434--1441. IEEE, 2010.

\bibitem{furukawa2009accurate}
Yasutaka Furukawa and Jean Ponce.
\newblock Accurate, dense, and robust multiview stereopsis.
\newblock {\em IEEE transactions on pattern analysis and machine intelligence},
  32(8):1362--1376, 2009.

\bibitem{galliani2015massively}
Silvano Galliani, Katrin Lasinger, and Konrad Schindler.
\newblock Massively parallel multiview stereopsis by surface normal diffusion.
\newblock In {\em Proceedings of the IEEE International Conference on Computer
  Vision}, pages 873--881, 2015.

\bibitem{giang2021curvature}
Khang~Truong Giang, Soohwan Song, and Sungho Jo.
\newblock Curvature-guided dynamic scale networks for multi-view stereo.
\newblock {\em arXiv preprint arXiv:2112.05999}, 2021.

\bibitem{goesele2007multi}
Michael Goesele, Noah Snavely, Brian Curless, Hugues Hoppe, and Steven~M Seitz.
\newblock Multi-view stereo for community photo collections.
\newblock In {\em 2007 IEEE 11th International Conference on Computer Vision},
  pages 1--8. IEEE, 2007.

\bibitem{casmvsnet}
Xiaodong Gu, Zhiwen Fan, Siyu Zhu, Zuozhuo Dai, Feitong Tan, and Ping Tan.
\newblock Cascade cost volume for high-resolution multi-view stereo and stereo
  matching.
\newblock 2019.

\bibitem{gu2020cascade}
Xiaodong Gu, Zhiwen Fan, Siyu Zhu, Zuozhuo Dai, Feitong Tan, and Ping Tan.
\newblock Cascade cost volume for high-resolution multi-view stereo and stereo
  matching.
\newblock In {\em Proceedings of the IEEE/CVF conference on computer vision and
  pattern recognition}, pages 2495--2504, 2020.

\bibitem{resnet}
Kaiming He, Xiangyu Zhang, Shaoqing Ren, and Jian Sun.
\newblock Deep residual learning for image recognition.
\newblock In {\em Proceedings of the IEEE conference on computer vision and
  pattern recognition}, pages 770--778, 2016.

\bibitem{hernandez2008multiview}
Carlos Hernandez, George Vogiatzis, and Roberto Cipolla.
\newblock Multiview photometric stereo.
\newblock {\em IEEE Transactions on Pattern Analysis and Machine Intelligence},
  30(3):548--554, 2008.

\bibitem{huang2018deepmvs}
Po-Han Huang, Kevin Matzen, Johannes Kopf, Narendra Ahuja, and Jia-Bin Huang.
\newblock Deepmvs: Learning multi-view stereopsis.
\newblock In {\em Proceedings of the IEEE Conference on Computer Vision and
  Pattern Recognition}, pages 2821--2830, 2018.

\bibitem{pstransformer}
Satoshi Ikehata.
\newblock Ps-transformer: Learning sparse photometric stereo network using
  self-attention mechanism.
\newblock {\em arXiv preprint arXiv:2211.11386}, 2022.

\bibitem{ji2017surfacenet}
Mengqi Ji, Juergen Gall, Haitian Zheng, Yebin Liu, and Lu Fang.
\newblock Surfacenet: An end-to-end 3d neural network for multiview stereopsis.
\newblock In {\em Proceedings of the IEEE International Conference on Computer
  Vision}, pages 2307--2315, 2017.

\bibitem{kang2001handling}
Sing~Bing Kang, Richard Szeliski, and Jinxiang Chai.
\newblock Handling occlusions in dense multi-view stereo.
\newblock In {\em Proceedings of the 2001 IEEE Computer Society Conference on
  Computer Vision and Pattern Recognition. CVPR 2001}, volume~1, pages I--I.
  IEEE, 2001.

\bibitem{kaya2022uncertainty}
Berk Kaya, Suryansh Kumar, Carlos Oliveira, Vittorio Ferrari, and Luc Van~Gool.
\newblock Uncertainty-aware deep multi-view photometric stereo.
\newblock In {\em Proceedings of the IEEE/CVF Conference on Computer Vision and
  Pattern Recognition}, pages 12601--12611, 2022.

\bibitem{kaya2023multi}
Berk Kaya, Suryansh Kumar, Carlos Oliveira, Vittorio Ferrari, and Luc Van~Gool.
\newblock Multi-view photometric stereo revisited.
\newblock In {\em Proceedings of the IEEE/CVF Winter Conference on Applications
  of Computer Vision}, pages 3126--3135, 2023.

\bibitem{kaya2022neural}
Berk Kaya, Suryansh Kumar, Francesco Sarno, Vittorio Ferrari, and Luc Van~Gool.
\newblock Neural radiance fields approach to deep multi-view photometric
  stereo.
\newblock In {\em Proceedings of the IEEE/CVF Winter Conference on Applications
  of Computer Vision}, pages 1965--1977, 2022.

\bibitem{kazhdan2006poisson}
Michael Kazhdan, Matthew Bolitho, and Hugues Hoppe.
\newblock Poisson surface reconstruction.
\newblock In {\em Proceedings of the fourth Eurographics symposium on Geometry
  processing}, volume~7, page~0, 2006.

\bibitem{kazhdan2013screened}
Michael Kazhdan and Hugues Hoppe.
\newblock Screened poisson surface reconstruction.
\newblock {\em ACM Transactions on Graphics (ToG)}, 32(3):1--13, 2013.

\bibitem{kingma2014adam}
Diederik~P Kingma and Jimmy Ba.
\newblock Adam: A method for stochastic optimization.
\newblock {\em arXiv preprint arXiv:1412.6980}, 2014.

\bibitem{knapitsch2017tanks}
Arno Knapitsch, Jaesik Park, Qian-Yi Zhou, and Vladlen Koltun.
\newblock Tanks and temples: Benchmarking large-scale scene reconstruction.
\newblock {\em ACM Transactions on Graphics (ToG)}, 36(4):1--13, 2017.

\bibitem{lassner2020pulsar}
Christoph Lassner and Michael Zollh\"ofer.
\newblock Pulsar: Efficient sphere-based neural rendering.
\newblock {\em arXiv:2004.07484}, 2020.

\bibitem{li2020multi}
Min Li, Zhenglong Zhou, Zhe Wu, Boxin Shi, Changyu Diao, and Ping Tan.
\newblock Multi-view photometric stereo: A robust solution and benchmark
  dataset for spatially varying isotropic materials.
\newblock {\em IEEE Transactions on Image Processing}, 29:4159--4173, 2020.

\bibitem{li2018learning}
Zhengqin Li, Zexiang Xu, Ravi Ramamoorthi, Kalyan Sunkavalli, and Manmohan
  Chandraker.
\newblock Learning to reconstruct shape and spatially-varying reflectance from
  a single image.
\newblock In {\em SIGGRAPH Asia 2018 Technical Papers}, page 269. ACM, 2018.

\bibitem{lichy2022fast}
Daniel Lichy, Soumyadip Sengupta, and David~W Jacobs.
\newblock Fast light-weight near-field photometric stereo.
\newblock In {\em Proceedings of the IEEE/CVF Conference on Computer Vision and
  Pattern Recognition}, pages 12612--12621, 2022.

\bibitem{Lichy_2022_CVPR}
Daniel Lichy, Soumyadip Sengupta, and David~W. Jacobs.
\newblock Fast light-weight near-field photometric stereo.
\newblock In {\em Proceedings of the IEEE/CVF Conference on Computer Vision and
  Pattern Recognition (CVPR)}, pages 12612--12621, June 2022.

\bibitem{Lichy_2021_CVPR}
Daniel Lichy, Jiaye Wu, Soumyadip Sengupta, and David~W. Jacobs.
\newblock Shape and material capture at home.
\newblock In {\em Proceedings of the IEEE/CVF Conference on Computer Vision and
  Pattern Recognition (CVPR)}, pages 6123--6133, June 2021.

\bibitem{lichy2021shape}
Daniel Lichy, Jiaye Wu, Soumyadip Sengupta, and David~W Jacobs.
\newblock Shape and material capture at home.
\newblock In {\em Proceedings of the IEEE/CVF Conference on Computer Vision and
  Pattern Recognition}, pages 6123--6133, 2021.

\bibitem{marching_cube}
William~E Lorensen and Harvey~E Cline.
\newblock Marching cubes: A high resolution 3d surface construction algorithm.
\newblock {\em ACM siggraph computer graphics}, 21(4):163--169, 1987.

\bibitem{luo2019p}
Keyang Luo, Tao Guan, Lili Ju, Haipeng Huang, and Yawei Luo.
\newblock P-mvsnet: Learning patch-wise matching confidence aggregation for
  multi-view stereo.
\newblock In {\em Proceedings of the IEEE/CVF International Conference on
  Computer Vision}, pages 10452--10461, 2019.

\bibitem{mescheder2019occupancy}
Lars Mescheder, Michael Oechsle, Michael Niemeyer, Sebastian Nowozin, and
  Andreas Geiger.
\newblock Occupancy networks: Learning 3d reconstruction in function space.
\newblock In {\em Proceedings of the IEEE/CVF conference on computer vision and
  pattern recognition}, pages 4460--4470, 2019.

\bibitem{mildenhall2021nerf}
Ben Mildenhall, Pratul~P Srinivasan, Matthew Tancik, Jonathan~T Barron, Ravi
  Ramamoorthi, and Ren Ng.
\newblock Nerf: Representing scenes as neural radiance fields for view
  synthesis.
\newblock {\em Communications of the ACM}, 65(1):99--106, 2021.

\bibitem{nehab2005efficiently}
Diego Nehab, Szymon Rusinkiewicz, James Davis, and Ravi Ramamoorthi.
\newblock Efficiently combining positions and normals for precise 3d geometry.
\newblock {\em ACM transactions on graphics (TOG)}, 24(3):536--543, 2005.

\bibitem{oechsle2021unisurf}
Michael Oechsle, Songyou Peng, and Andreas Geiger.
\newblock Unisurf: Unifying neural implicit surfaces and radiance fields for
  multi-view reconstruction.
\newblock In {\em Proceedings of the IEEE/CVF International Conference on
  Computer Vision}, pages 5589--5599, 2021.

\bibitem{park2013multiview}
Jaesik Park, Sudipta~N Sinha, Yasuyuki Matsushita, Yu-Wing Tai, and In~So
  Kweon.
\newblock Multiview photometric stereo using planar mesh parameterization.
\newblock In {\em Proceedings of the IEEE International Conference on Computer
  Vision}, pages 1161--1168, 2013.

\bibitem{park2016robust}
Jaesik Park, Sudipta~N Sinha, Yasuyuki Matsushita, Yu-Wing Tai, and In~So
  Kweon.
\newblock Robust multiview photometric stereo using planar mesh
  parameterization.
\newblock {\em IEEE transactions on pattern analysis and machine intelligence},
  39(8):1591--1604, 2016.

\bibitem{paszke2019pytorch}
Adam Paszke, Sam Gross, Francisco Massa, Adam Lerer, James Bradbury, Gregory
  Chanan, Trevor Killeen, Zeming Lin, Natalia Gimelshein, Luca Antiga, et~al.
\newblock Pytorch: An imperative style, high-performance deep learning library.
\newblock {\em Advances in neural information processing systems}, 32, 2019.

\bibitem{ravi2020pytorch3d}
Nikhila Ravi, Jeremy Reizenstein, David Novotny, Taylor Gordon, Wan-Yen Lo,
  Justin Johnson, and Georgia Gkioxari.
\newblock Accelerating 3d deep learning with pytorch3d.
\newblock {\em arXiv:2007.08501}, 2020.

\bibitem{schonberger2016pixelwise}
Johannes~L Sch{\"o}nberger, Enliang Zheng, Jan-Michael Frahm, and Marc
  Pollefeys.
\newblock Pixelwise view selection for unstructured multi-view stereo.
\newblock In {\em Computer Vision--ECCV 2016: 14th European Conference,
  Amsterdam, The Netherlands, October 11-14, 2016, Proceedings, Part III 14},
  pages 501--518. Springer, 2016.

\bibitem{sengupta2018sfsnet}
Soumyadip Sengupta, Angjoo Kanazawa, Carlos~D Castillo, and David~W Jacobs.
\newblock Sfsnet: Learning shape, reflectance and illuminance of facesin the
  wild'.
\newblock In {\em Proceedings of the IEEE Conference on Computer Vision and
  Pattern Recognition}, pages 6296--6305, 2018.

\bibitem{diligent_data}
Boxin Shi, Zhe~Wu Mo, Dinglong Duan, Sai-Kit Yeung, and Ping Tan.
\newblock A benchmark dataset and evalution for non-lambertian and uncalibrated
  photometric stereo.
\newblock {\em IEEE Trans. on Pattern Analysis and Machine Intelligence
  ({TPAMI})}, 41(2):271--284, 2019.

\bibitem{srinivasan2021nerv}
Pratul~P Srinivasan, Boyang Deng, Xiuming Zhang, Matthew Tancik, Ben
  Mildenhall, and Jonathan~T Barron.
\newblock Nerv: Neural reflectance and visibility fields for relighting and
  view synthesis.
\newblock In {\em Proceedings of the IEEE/CVF Conference on Computer Vision and
  Pattern Recognition}, pages 7495--7504, 2021.

\bibitem{strecha2004wide}
Christoph Strecha, Rik Fransens, and Luc Van~Gool.
\newblock Wide-baseline stereo from multiple views: a probabilistic account.
\newblock In {\em Proceedings of the 2004 IEEE Computer Society Conference on
  Computer Vision and Pattern Recognition, 2004. CVPR 2004.}, volume~1, pages
  I--I. IEEE, 2004.

\bibitem{strecha2006combined}
Christoph Strecha, Rik Fransens, and Luc Van~Gool.
\newblock Combined depth and outlier estimation in multi-view stereo.
\newblock In {\em 2006 IEEE Computer Society Conference on Computer Vision and
  Pattern Recognition (CVPR'06)}, volume~2, pages 2394--2401. IEEE, 2006.

\bibitem{sun2021neucon}
Jiaming Sun, Yiming Xie, Linghao Chen, Xiaowei Zhou, and Hujun Bao.
\newblock {NeuralRecon}: Real-time coherent {3D} reconstruction from monocular
  video.
\newblock {\em CVPR}, 2021.

\bibitem{sculpture_data}
Olivia Wiles and Andrew Zisserman.
\newblock Silnet : Single- and multi-view reconstruction by learning from
  silhouettes.
\newblock In {\em British Machine Vision Conference 2017, {BMVC} 2017, London,
  UK, September 4-7, 2017}. {BMVA} Press, 2017.

\bibitem{woodham1980photometric}
Robert~J Woodham.
\newblock Photometric method for determining surface orientation from multiple
  images.
\newblock {\em Optical engineering}, 19(1):139--144, 1980.

\bibitem{xu2020learning}
Qingshan Xu and Wenbing Tao.
\newblock Learning inverse depth regression for multi-view stereo with
  correlation cost volume.
\newblock In {\em Proceedings of the AAAI Conference on Artificial
  Intelligence}, volume~34, pages 12508--12515, 2020.

\bibitem{xu2018deep}
Zexiang Xu, Kalyan Sunkavalli, Sunil Hadap, and Ravi Ramamoorthi.
\newblock Deep image-based relighting from optimal sparse samples.
\newblock {\em ACM Transactions on Graphics (TOG)}, 37(4):126, 2018.

\bibitem{yang2022ps}
Wenqi Yang, Guanying Chen, Chaofeng Chen, Zhenfang Chen, and Kwan-Yee~K Wong.
\newblock Ps-nerf: Neural inverse rendering for multi-view photometric stereo.
\newblock {\em arXiv preprint arXiv:2207.11406}, 2022.

\bibitem{yao2018mvsnet}
Yao Yao, Zixin Luo, Shiwei Li, Tian Fang, and Long Quan.
\newblock Mvsnet: Depth inference for unstructured multi-view stereo.
\newblock In {\em Proceedings of the European conference on computer vision
  (ECCV)}, pages 767--783, 2018.

\bibitem{yao2019recurrent}
Yao Yao, Zixin Luo, Shiwei Li, Tianwei Shen, Tian Fang, and Long Quan.
\newblock Recurrent mvsnet for high-resolution multi-view stereo depth
  inference.
\newblock {\em Computer Vision and Pattern Recognition (CVPR)}, 2019.

\bibitem{yariv2020multiview}
Lior Yariv, Yoni Kasten, Dror Moran, Meirav Galun, Matan Atzmon, Basri Ronen,
  and Yaron Lipman.
\newblock Multiview neural surface reconstruction by disentangling geometry and
  appearance.
\newblock {\em Advances in Neural Information Processing Systems},
  33:2492--2502, 2020.

\bibitem{pixelnerf}
Alex Yu, Vickie Ye, Matthew Tancik, and Angjoo Kanazawa.
\newblock pixelnerf: Neural radiance fields from one or few images, 2020.

\bibitem{zhang2021physg}
Kai Zhang, Fujun Luan, Qianqian Wang, Kavita Bala, and Noah Snavely.
\newblock Physg: Inverse rendering with spherical gaussians for physics-based
  material editing and relighting.
\newblock In {\em Proceedings of the IEEE/CVF Conference on Computer Vision and
  Pattern Recognition}, pages 5453--5462, 2021.

\bibitem{zhang2021nerfactor}
Xiuming Zhang, Pratul~P Srinivasan, Boyang Deng, Paul Debevec, William~T
  Freeman, and Jonathan~T Barron.
\newblock Nerfactor: Neural factorization of shape and reflectance under an
  unknown illumination.
\newblock {\em ACM Transactions on Graphics (TOG)}, 40(6):1--18, 2021.

\bibitem{zhang1994iterative}
Zhengyou Zhang.
\newblock Iterative point matching for registration of free-form curves and
  surfaces.
\newblock {\em International journal of computer vision}, 13(2):119--152, 1994.

\bibitem{zhou2013multi}
Zhenglong Zhou, Zhe Wu, and Ping Tan.
\newblock Multi-view photometric stereo with spatially varying isotropic
  materials.
\newblock In {\em Proceedings of the IEEE Conference on Computer Vision and
  Pattern Recognition}, pages 1482--1489, 2013.

\end{thebibliography}
}

\newpage

\section{Appendix}
\subsection{Overview}
    In this supplementary material, we will include the following contents:
    \begin{itemize}
        \item We describe more details about our \textbf{sMVPS dataset} in \textbf{Section 2} and show additional example images in Figure \ref{fig:supp_sMVPS_sculpture} and Figure \ref{fig:supp_sMVPS_random}.

        \item We provide additional \textbf{experiment details}, including notations we use for network architecture and implementation details in \textbf{Section 3}.

        \item We explain our \textbf{mesh extraction pipeline} in detail in \textbf{Section 4} together with the parameters we use.

        \item We provide the equations of the \textbf{evaluation metrics} we use in \textbf{Section 5}.

        \item In the main paper, we provide L1 Chamfer distance and F-score with L2 distance after ICP \cite{4767965, besl1992method, chen1992object, zhang1994iterative}. Here in \textbf{Section 6}, we also provide \textbf{results of L1 Chamfer distance and F-score with L2 distance before ICP} in Table \ref{table:supp_3d_results_chamfer} and \ref{table:supp_3d_results_f_score}.

        \item To further compare with generalizable MVS methods, in addtion to CasMVSNet \cite{casmvsnet} in the main paper, we also \textbf{compare our method with a state-of-the-art MVS method, TransMVSNet \cite{transmvsnet}}, in \textbf{Section 7}. TransMVSNet \cite{transmvsnet} is built upon CasMVSNet and adopts a transformer to consider intra-image and inter-image feature interactions, which further improves the result of CasMVSNet \cite{casmvsnet}. See Table \ref{table:supp_transmvsnet} for the comparison. 

        \item We include additional qualitative results. We show the global shape of reconstructed mesh from each method under three different views in Figure \ref{fig:bear} - \ref{fig:reading}. We also show additional zoomed areas for visual comparison between meshes in Figure \ref{fig:supp_zoomed}.

    \end{itemize}

\subsection{sMVPS datasets}

\textbf{Object and Camera Positioning}
For both synthetic datasets objects are placed at the center of world coordinates with the object’s up direction along the z-axis. Objects are scaled to be inside a sphere of radius one. We use a pinhole camera for rendering with an FOV of 9.3$^\circ$, which is similar to the FOV used to capture the DiligentMV dataset. Camera positions are most easily described in spherical coordinates i.e. an azimuth angle, a polar angle, and a radial distance. The azimuth angle for the $i$th camera is $(18 + X_i)^\circ$ Where $X_i$ is a uniform random number between -3 and 3, and $i$ runs from 0 to 19. The polar angle for each camera is sampled uniformly from 62-64$^\circ$. The radial distance is sampled uniformly between 14 and 16.5. This distance is chosen so the object occupies the majority of the image.

\textbf{Light Positioning}
Each view is rendered under 10 directional lights. The first light is always co-directional with the camera while the other 9 are randomly sampled from the spherical cap centered on the cameras optical axis with an angle of 45$^\circ$.

\textbf{BRDF}
To generate BRDFs we follow \cite{lichy2021shape}. Namely we use the Cook-Torrance BRDF model with spatially-varying albedo drawn from 415 free textures from \cite{3dtextures}, and randomly generated roughness as described in \cite{lichy2021shape}. Roughness is constant in the case of sMVPS-sculpture and constant for each primitive in the case of sMVPS-random.

\textbf{Object Meshes}
For the sMVPS-random dataset objects are drawn from the collections of random primitives generated by \cite{xu2018deep}  using a 90-10 train/test split. For the sMVPS-sculpture dataset we use the following meshes from \cite{sculpture_data} to render the train set: nymphe-seated, standing-isis-priest, the-slave-girl, thor, three-danish-polar-explorers, tiger-devouring-a-gavial, two-wrestlers-in-combat,ugolino-and-his-sons, virgin-and-child, woman-associated-with-the-cult-of-isis, wounded-amazon, wounded-cupid,wrestling-decimated-cleaned and the mesh virgin-mary-with-her-dead-son for the test set.

\textbf{Rendering}
Images are rendered with Mitsuba 2 using the path-tracer integration method. We render at a resolution of 612x512 with 128 samples-per-pixel.

\begin{figure*}[t]
    \begin{center}
    \includegraphics[width=\linewidth]{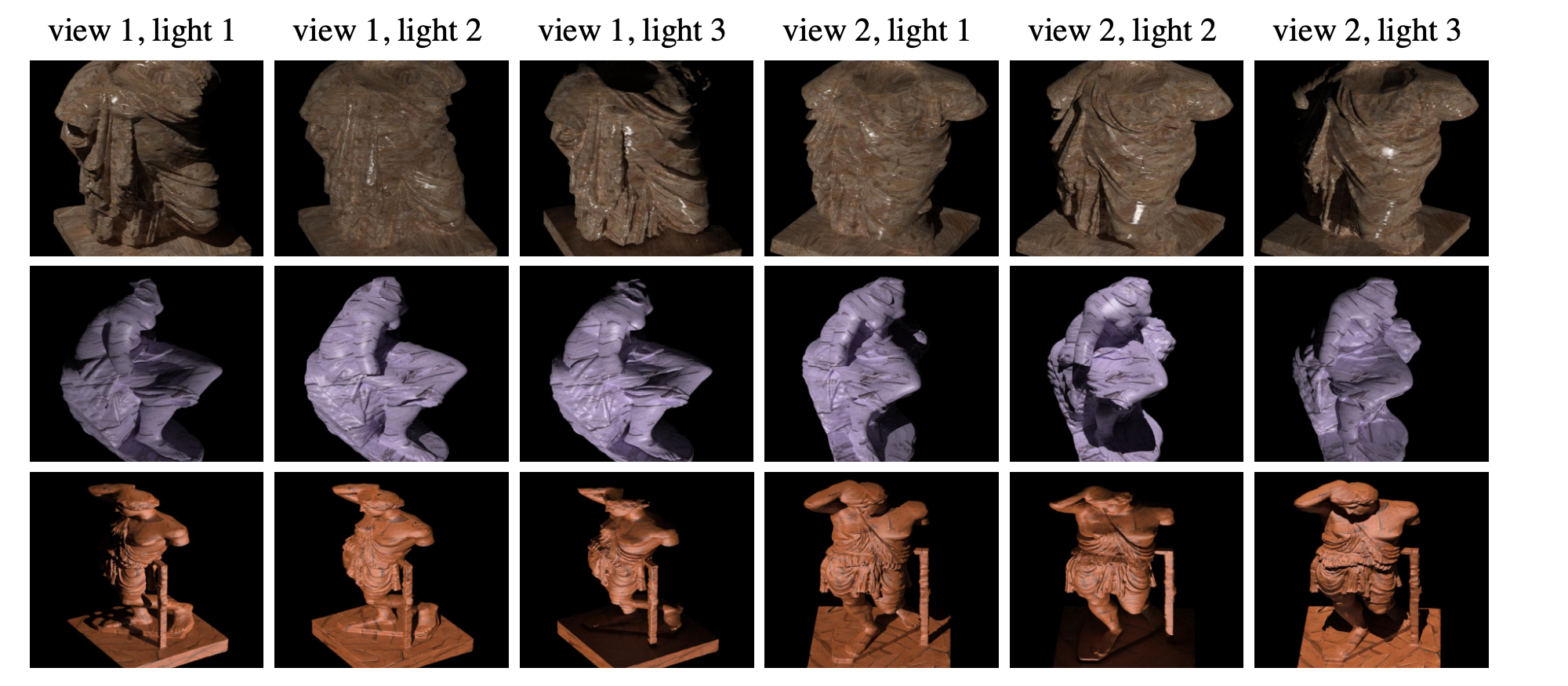}
    \end{center}
       \caption{Additional example images of sMVPS-sculpture.}
    \label{fig:supp_sMVPS_sculpture}
    \end{figure*}

\begin{figure*}
    \begin{center}
    \includegraphics[width=\linewidth]{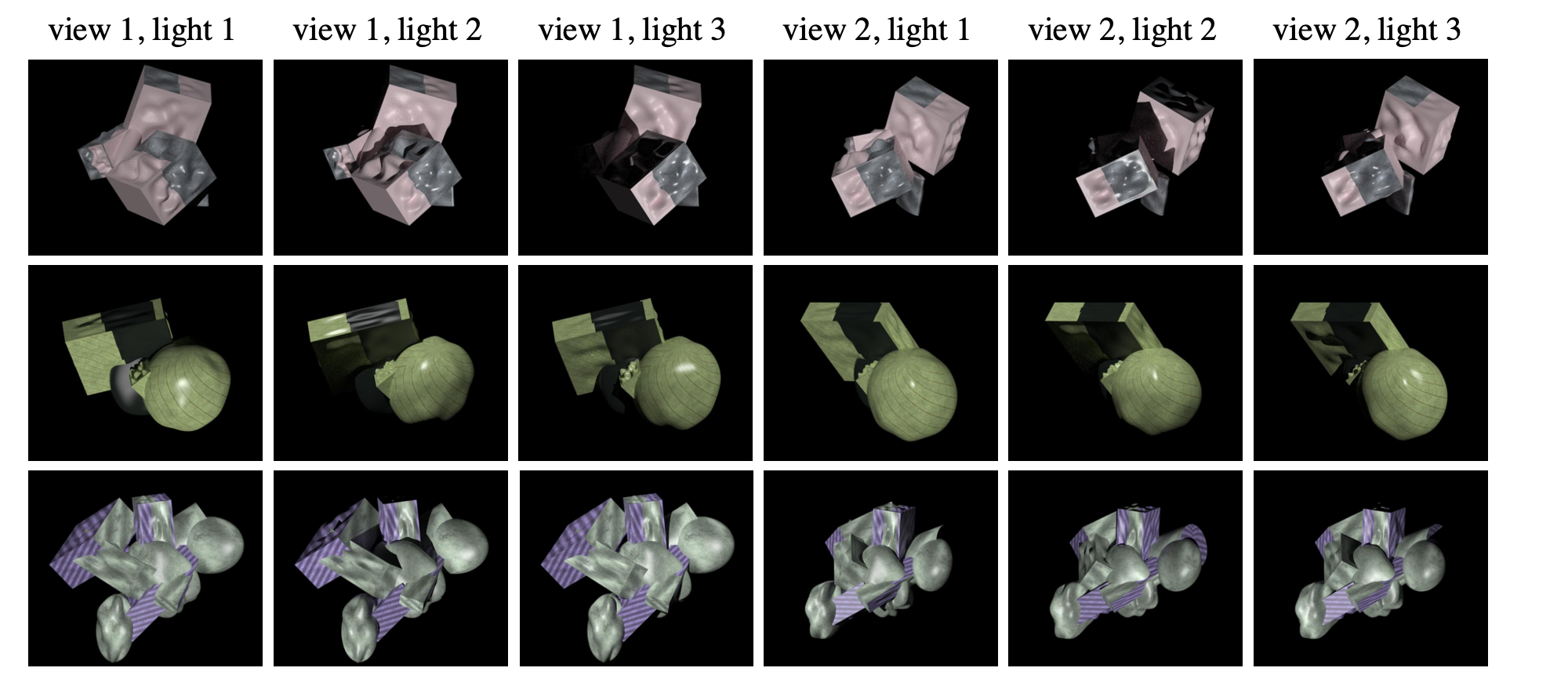}
    \end{center}
       \caption{Additional example images of sMVPS-random.}
    \label{fig:supp_sMVPS_random}
    \end{figure*}

\begin{figure*}[t]
    \begin{center}
    \includegraphics[width=0.95\linewidth]{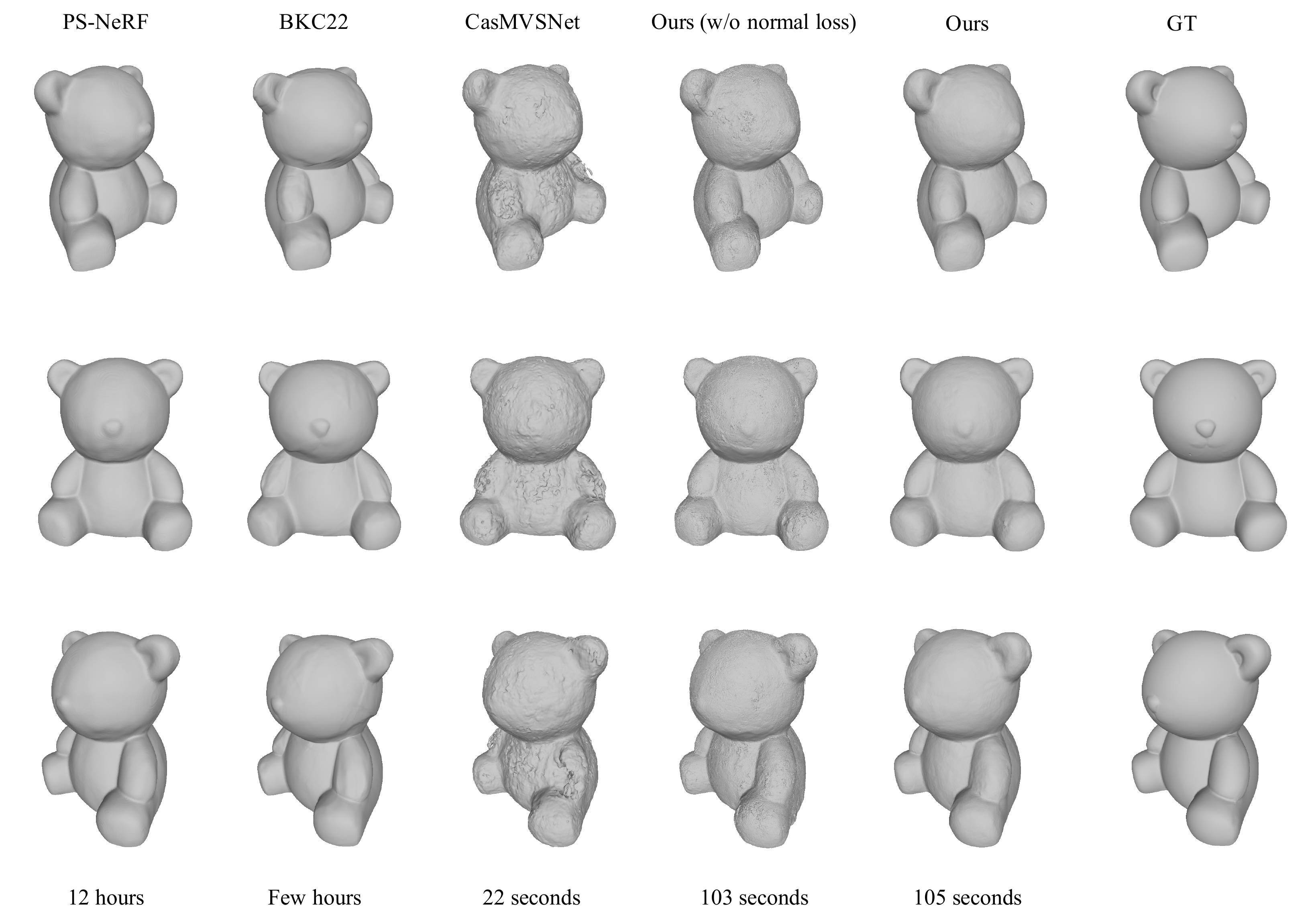}
    \end{center}
       \caption{Reconstruction of BEAR under three different views (left-side, front, right-side) in DiLiGenT-MV \cite{li2020multi}. Last row is reconstruction time.}
    \label{fig:bear}
    \end{figure*}

\begin{figure*}[t]
    \begin{center}
    \includegraphics[width=0.95\linewidth]{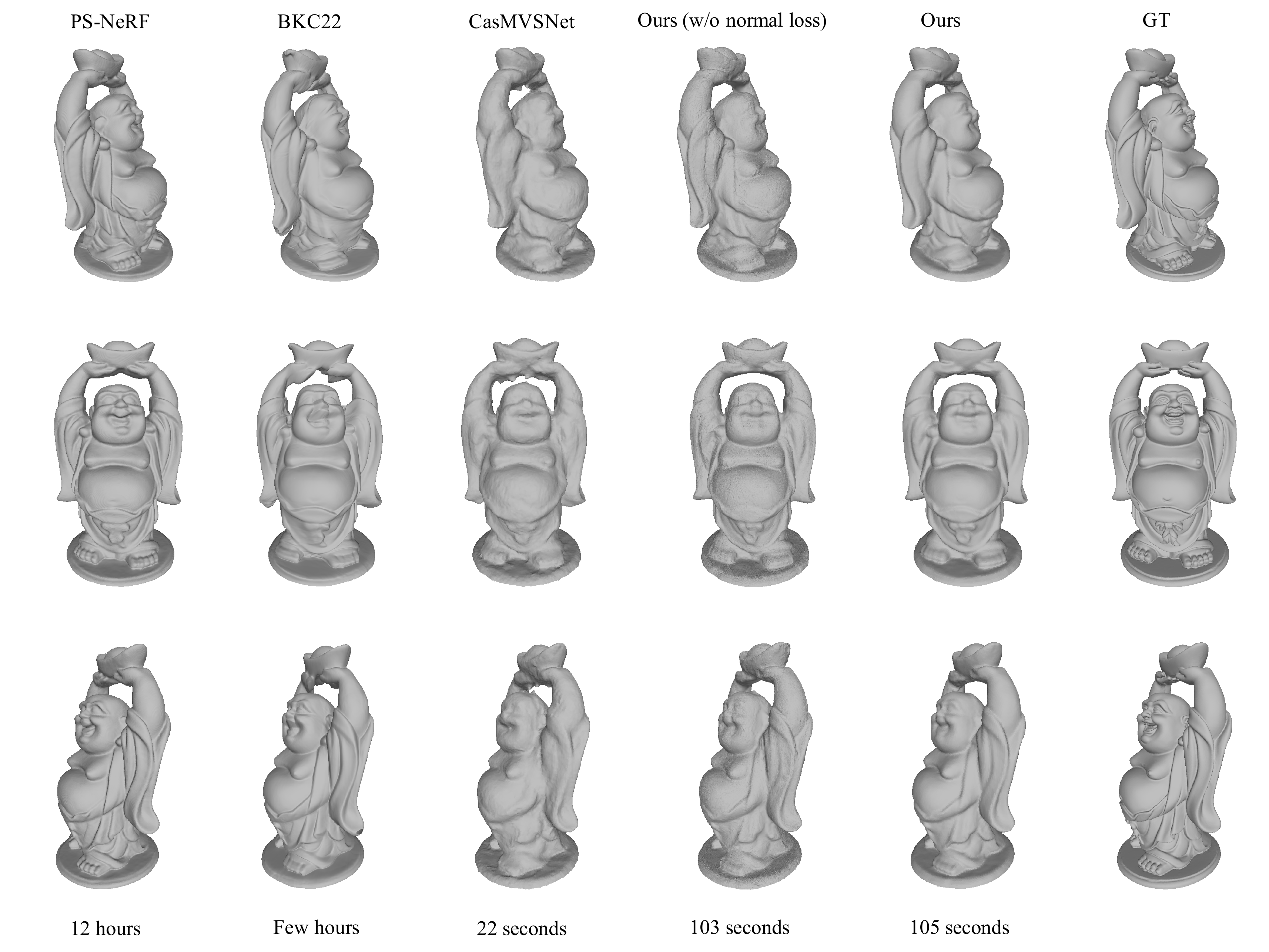}
    \end{center}
       \caption{Reconstruction of BUDDHA under three different views (left-side, front, right-side) in DiLiGenT-MV \cite{li2020multi}. Last row is reconstruction time.}
    \label{fig:buddha}
    \end{figure*}

\begin{figure*}[t]
    \begin{center}
    \includegraphics[width=0.95\linewidth]{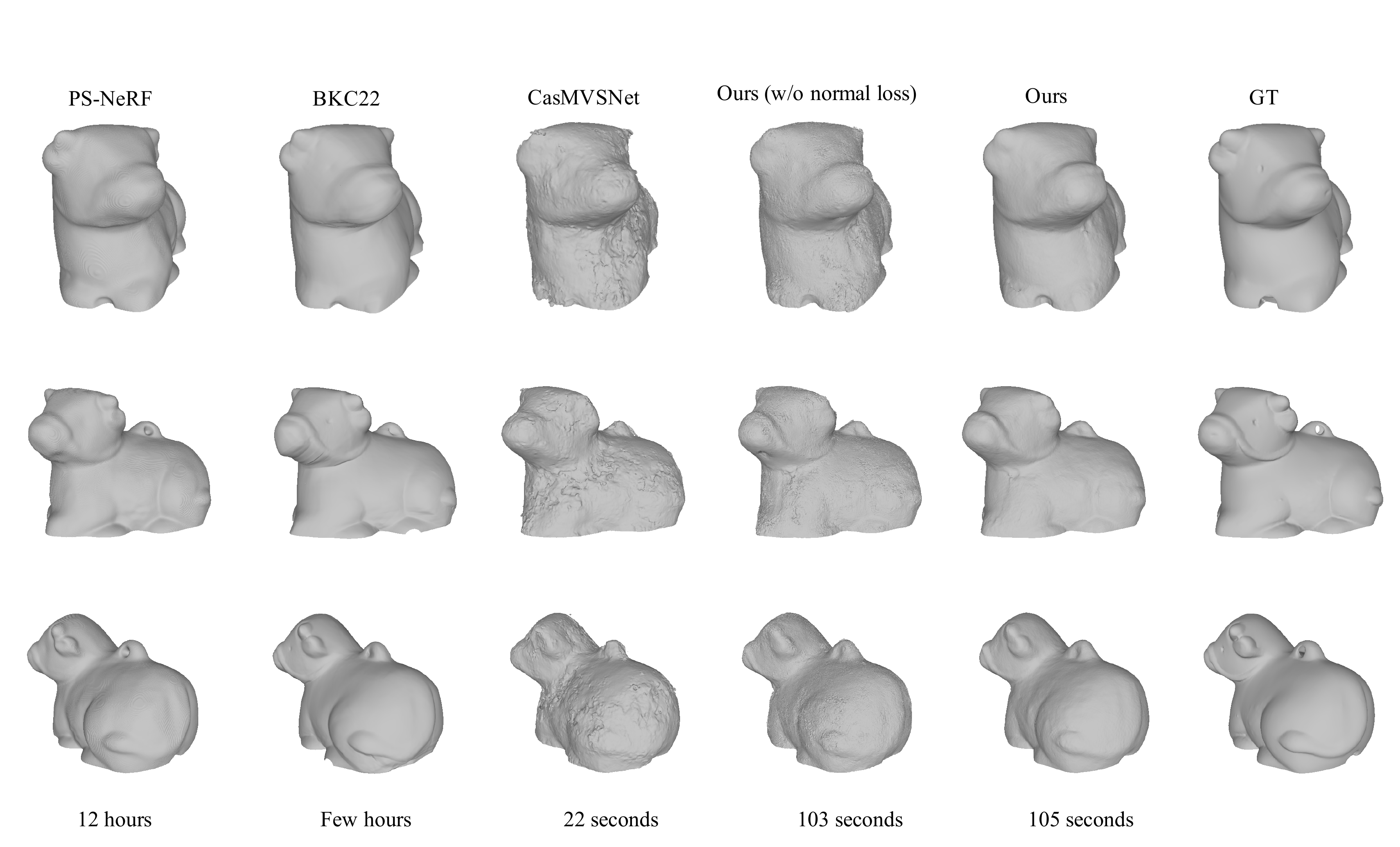}
    \end{center}
       \caption{Reconstruction of COW under three different views (front, right-side, back) in DiLiGenT-MV \cite{li2020multi}. Last row is reconstruction time.}
    \label{fig:cow}
    \end{figure*}

\begin{figure*}[t]
    \begin{center}
    \includegraphics[width=0.95\linewidth]{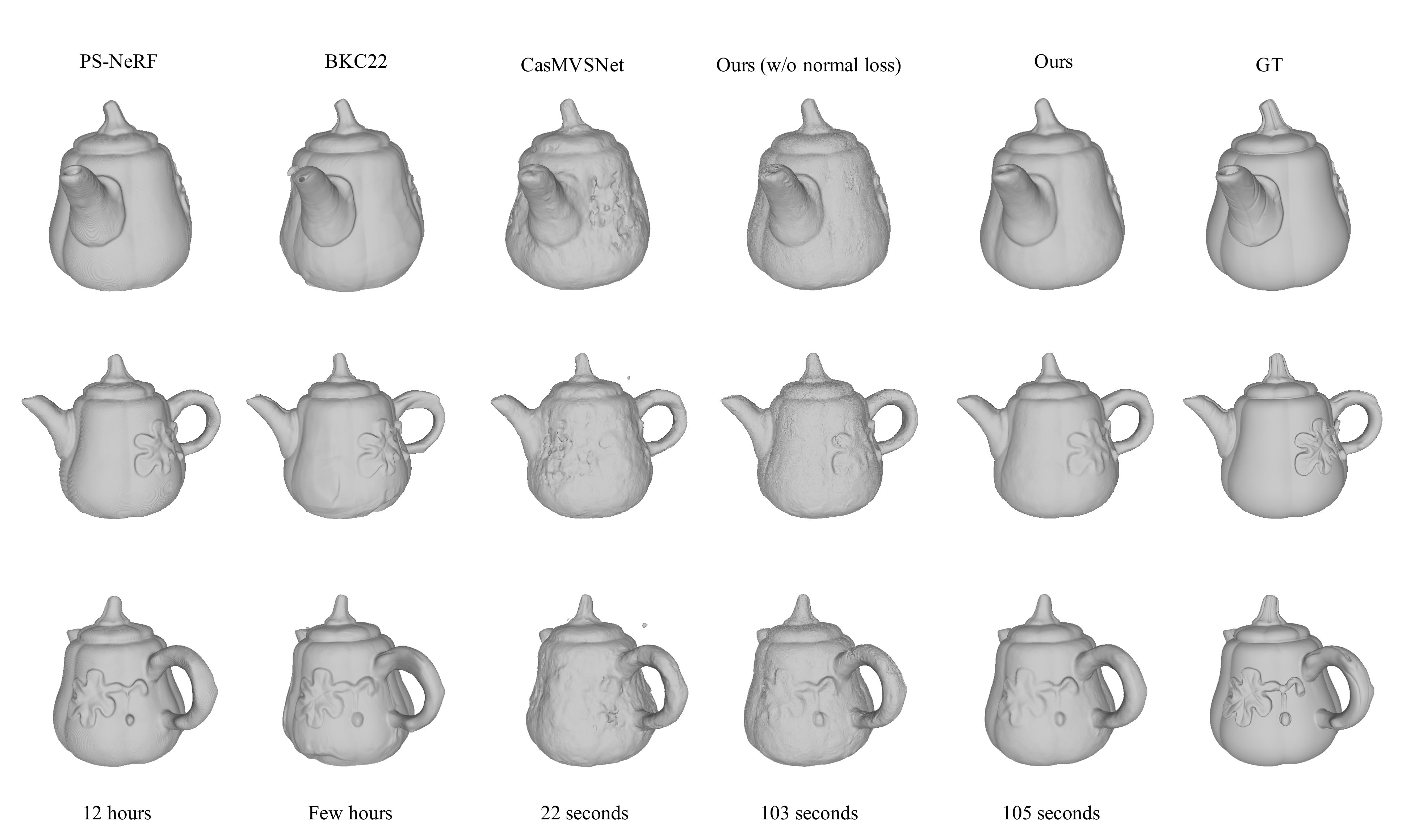}
    \end{center}
       \caption{Reconstruction of POT2 under three different views (left-side, front, right-side) in DiLiGenT-MV \cite{li2020multi}. Last row is reconstruction time.}
    \label{fig:pot2}
    \end{figure*}

\begin{figure*}[t]
    \begin{center}
    \includegraphics[width=0.95\linewidth]{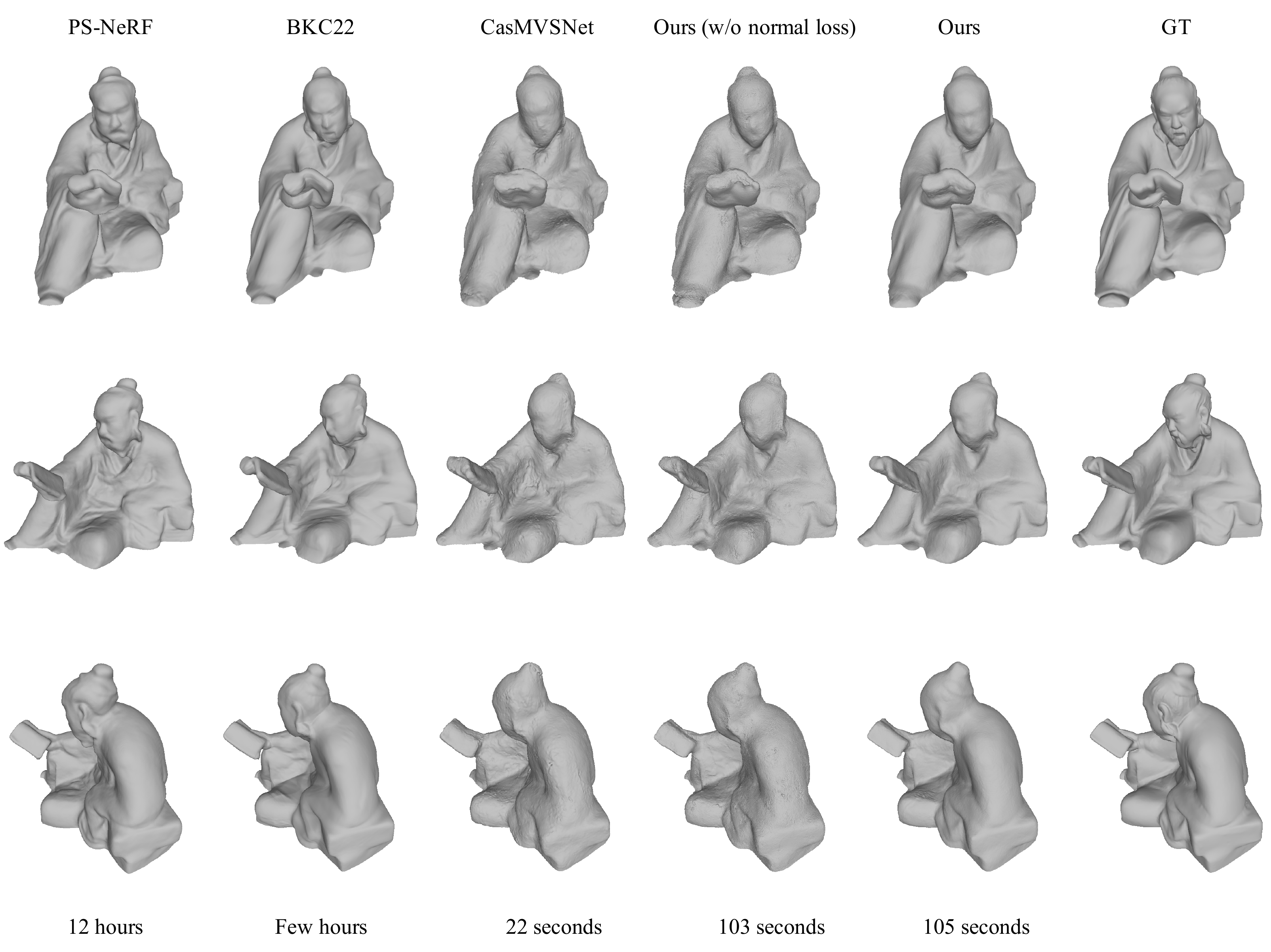}
    \end{center}
       \caption{Reconstruction of READING under three different views (front, right-side, back) in DiLiGenT-MV \cite{li2020multi}. Last row is reconstruction time.}
    \label{fig:reading}
    \end{figure*}

\begin{figure*}[t]
    \begin{center}
    \includegraphics[width=0.95\linewidth]{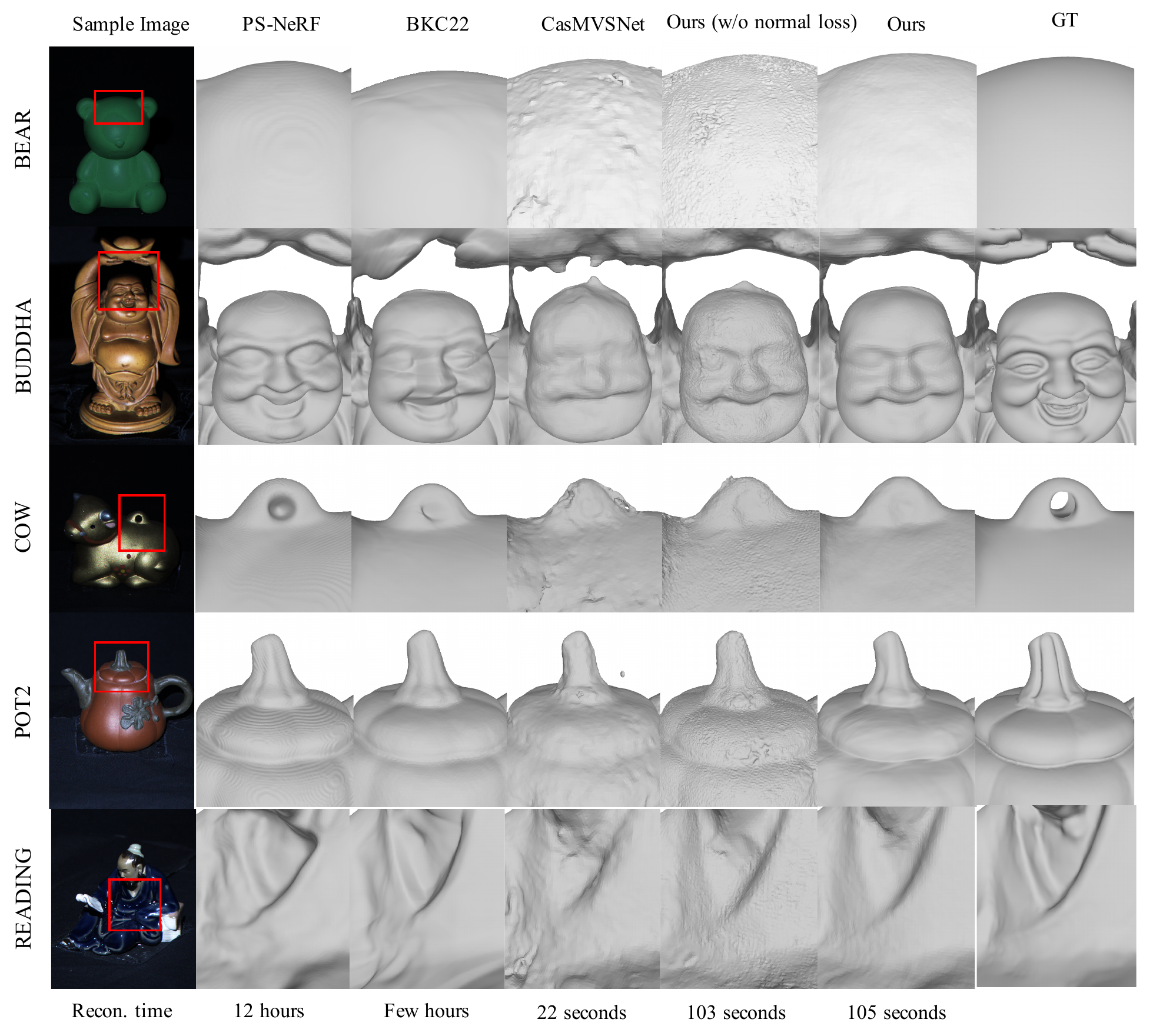}
    \end{center}
       \caption{Zoomed-in areas on meshes from all methods. We observe that in general PS-NeRF \cite{yang2022ps} provides mesh with global fine details while it also contains iso-contour pattern artifacts. Our method can provide smooth mesh with correct global shape even though it takes very short time compared with other methods.}
    \label{fig:supp_zoomed}
    \end{figure*}

\textbf{More Examples}
    To further show the diversity on surface shape, texture and material of our sMVPS datasets, we provide additional example images of sMVPS-sculpture in Figure \ref{fig:supp_sMVPS_sculpture} and sMVPS-random in Figure \ref{fig:supp_sMVPS_random}.

\subsection{Additional experiment details}

\subsubsection{Notation in Figure 1 of main paper}
    The architecture of our network is illustrated in Figure 1 in the paper and we describe the a few details and notations we use:
    
    \noindent \textbf{ResBlk:} Resnet block. It consists of \textit{conv2d(kernel=3) $\rightarrow$ BatchNorm $\rightarrow$ ReLu $\rightarrow$ conv2d(kernel=3) $\rightarrow$ BatchNorm}. And the input of this block is added to the output of this block as a residual connection \cite{resnet}.
    
    \noindent \textbf{Tconv:} ConvTranspose2d layer in Pytorch with kernel=3.

\subsubsection{Implementation details}
    Our model is implemented in Pytorch \cite{paszke2019pytorch} and we use a NVIDIA RTX A6000 GPU to train it. For input images, we crop them to $512 \times 512$ and rescale the pixel values to (0, 1). For each training sample, we use 3 views and 3 lightings. It is challenging to find correspondences for view selection in textureless regions, so we simply take the two adjacent views of a reference view as source views. To make our model more robust to different lighting configurations, we randomly sample 3 lightings and use the same lightings for all views, resulting in $3\times3=9$ images for each training sample. We use Adam \cite{kingma2014adam} optimizer and set betas as $(0.9, 0.999)$. We trained 50 epochs in total. The initial learning rate is 0.001 and it decays to half at steps [8, 12, 30, 40]. To get ground truth depth map of DiLiGent-MV \cite{li2020multi}, we render depth map from ground truth mesh and camera parameters.

\subsection{Mesh extraction pipeline}
    We use the same mesh extraction pipeline to recover 3D mesh from predicted depth maps for CasMVSNet \cite{casmvsnet}, Ours and TransMVSNet \cite{transmvsnet} for a fair comparison.
\subsubsection{Depth filtering}
    We use two kinds of masks to filter predicted depth maps. First, we employ 2D object mask to rule out background. This is because our model is only trained on pixels within an object. 
    Second, we apply geometric filtering to only keep depth predictions that are consistent across adjacent views. 
    For each object pixel in reference view, $p_{o}$, we have a predicted depth aligned with this view, $d_{o}$. 
    We lift $p_{o}$ to a 3D point, $P_{o}$, and project $P_{o}$ to a source view pixel, $p_{s}$. 
    Assume the predicted depth of source view at $p_{s}$ is $d_{s}$.
    By lifting $p_{s}$ using $d_{s}$, we get a 3D point $P_{s}$. 
    Projecting $P_{s}$ back to reference view results in a reprojected pixel, $p_{r}$ and a depth $d_{r}$. 
    We set thresholds for the distance between the original pixel, $p_{o}$, and reprojected pixel, $p_{r}$, as well as relative difference between $d_{o}$ and $d_{r}$ as follows:
    \begin{align}
        dist(p_{o}, p_{r}) < 1 , \\
        abs(d_{o} - d_{r}) / d_{o} < 0.01
    \end{align}
    For $p_{o}$ and $d_{o}$, 
    we check its geometric consistency with each source view and keep it only if it is consistent with at least one source view.

\subsubsection{Depth fusion}
    After the depth filtering step, we combine each depth map in a fusion step. For an object pixel $p_{o}$, we simply average over $d_{o}$ and  all the estimations from source views that are consistent with it, $d_{si}$ for $i=1, ..., i_{N}$, where $i_{N}$ is the total number of geometric consistent neighboring views, and use this average as depth at $p_{o}$. 
    We then lift $p_{o}$ to a vertex in point cloud and attach the predicted normal, $n_{o}$, to it. This way, we get point cloud utilizing information from depth maps of all views.

    Note there are other possible depth fusion methods, \eg GIPUMA \cite{galliani2015massively}, some of which may achieve better fusion performance for certain datasets. But there is no method that is better for all datasets, so we leave exploration in this direction as a future work.

\subsubsection{Surface reconstruction}
    We apply Screend Poisson Surface Reconstruction (SPSR) \cite{kazhdan2013screened} to recover mesh from point cloud. We use same set of parameters for all methods and all objects. Specifically, we set $reconstruction\_depth=8$, $minimum\_number\_of\_samples=1.5$ and $interpolation\_weight=4$. 
    Note that before recovering surfaces, an extra step of computing normal based on the point cloud is needed for methods without normal prediction, i.e., CasMVSNet \cite{casmvsnet} and TransMVSNet \cite{transmvsnet}.

\subsubsection{Evaluation metrics details}
    We use L1 Chamfer distance and F-socre with L2 distance (threshold at 1mm) to evaluate the quality of reconstructed mesh. Both metrics are applied to two sets of 3D points, which are vertices of reconstructed mesh and ground truth mesh. 
    
    Give two point sets, $\mathcal{R}$ and $\mathcal{G}$, L1 Chamfer distance is defined as follows:
    \begin{align}
        CD(\mathcal{R}, \mathcal{G})=\frac{1}{|\mathcal{R}|}\sum_{x\in\mathcal{R}}\min_{y\in \mathcal{G}}\|x-y\|+\frac{1}{|\mathcal{G}|}\sum_{y\in\mathcal{G}}\min_{x\in \mathcal{R}}\|x-y\|.
    \end{align}
    
    We use the F-score similarly defined as \cite{knapitsch2017tanks}, for a reconstructed point $r \in \mathcal{R}$, its L2 distance to the ground truth mesh $\mathcal{G}$ is 
    \begin{align}
        e_{r\rightarrow \mathcal{G}}=\min_{g\in\mathcal{G}}\|r-g\|_2,
    \end{align}
    and for a ground truth point $g \in \mathcal{G}$, its distance to the reconstructed mesh is defined as:
    \begin{align}
        e_{g\rightarrow \mathcal{R}}=\min_{r\in\mathcal{R}}\|r-g\|_2,
    \end{align}
    
    The precision and recall for a threshold $d$ are:
    \begin{align}
        P(d) = \frac{1}{|\mathcal{R|}}\sum_{r\in\mathcal{R}}[e_{r\rightarrow\mathcal{G}}<d] \\
        R(d) = \frac{1}{|\mathcal{G}|}\sum_{g\in\mathcal{G}}[e_{g\rightarrow\mathcal{R}}<d]
    \end{align}

    F-score is the harmonic mean of precision and recall as a summary measure:
    \begin{align}
        F(d) = \frac{2P(d)R(d)}{P(d)+R(d)}
    \end{align}

\subsection{Additional results without ICPs}
    
    In the paper, we report results after ICP \cite{4767965, besl1992method, chen1992object, zhang1994iterative}, which is an extra registration step we applied to all meshes after being reconstructed. It is initially aimed to fairly compare our mesh with others as several methods indicate that they did registration after extracting meshes \cite{li2020multi, kaya2022neural}. However, we find it also helpful to improve accuracy of other methods, even for those that already have registration applied. Since there is no standard way to do registration among existing methods, we applied ICP to meshes from all methods, regardless of whether they have done registration or not. 
    
    For a complete comparison, we also provide the quantitative results of L1 Chamfer distance and F-score with L2 distance (threshold at 1mm) without ICP \cite{4767965, besl1992method, chen1992object, zhang1994iterative} in Table \ref{table:supp_3d_results_chamfer} and Table \ref{table:supp_3d_results_f_score}, respectively. They show that even without registration, our method can still perform comparably with state-of-the-art methods with registration.

        \begin{table*}[t]
        \footnotesize
        \begin{center}
        \begin{tabular}{l|P{3em}P{3em}|P{3em}P{3em}P{4em}|c|c|P{7em}}
        \toprule
         &  \multicolumn{5}{c|}{{\small \textbf{Per-scene optimization}}} & \multicolumn{3}{c}{{\small \textbf{Generalizable}}}\\ 
        \hline
        Category  &  \multicolumn{2}{c|}{Manual Effort} & \multicolumn{3}{c|}{Standalone} & Single-view PS & MVS & MVPS\\ 
        \hline
        Method  & PJ16\cite{park2016robust} & LZ20\cite{li2020multi} & BKW22\cite{kaya2022neural} & BKC22\cite{kaya2022uncertainty} & PS-NeRF\cite{yang2022ps} & PS-Transformer\cite{pstransformer} &  CasMVSNet\cite{gu2020cascade}- RT & Ours\\
        \hline 
        BEAR & 2.63 & 0.74 & 1.03 & 1.09 & \textbf{0.81} & 3.25 & 1.38 & \underline{0.91} \\
        BUDDHA & 1.18 & 0.99 & 2.44 & 1.19 & \textbf{0.98} & 4.44 & 1.30 & \underline{1.12} \\
        COW & 1.16 & 0.39 & 1.08 & 0.86 & \textbf{0.78} & 2.67 & 1.26 & \underline{0.80} \\
        POT2 & 3.27 & 0.69 & 1.32 & 1.32 & \textbf{0.81} & 2.92 & 1.43 & \underline{0.94} \\
        READING & 1.49 & 0.74 & 1.94 & 0.93 & 0.98 & 3.69 & \underline{0.83} & \textbf{0.76} \\
        \hline
        AVERAGE & 1.95 & 0.71 & 1.56 & 1.08 & \textbf{0.87} & 3.39 & 1.24 & \underline{0.91} \\
        \bottomrule
        \end{tabular}
        \end{center}
        \caption{L1 Chamfer Distance (lower is better) between reconstructed mesh and GT without ICP. `-RT' denotes trained on our synthetic MVPS dataset. For non-manual methods, the best result is shown in bold, 2nd best as underline. LZ20 \& PJ16 involve carefully crafted steps, manual efforts in finding correspondence, and an initial mesh or point cloud.}
        \label{table:supp_3d_results_chamfer}
        \end{table*}

        \begin{table*}[t]
        \footnotesize
        \begin{center}
        \begin{tabular}{l|P{2em}P{2em}|P{3em}P{3em}P{3em}P{4em}|P{7em}|P{6em}|P{7em}}
        \toprule
         &  \multicolumn{6}{c|}{{\small \textbf{Per-scene optimization}}} & \multicolumn{3}{c}{{\small \textbf{Generalizable}}}\\ 
        \hline
        Category  &  \multicolumn{2}{c|}{Manual Effort} & \multicolumn{4}{c|}{Standalone} & Single-view PS & MVS & MVPS\\ 
        \hline
        Method  & PJ16\cite{park2016robust} & LZ20\cite{li2020multi} & BKW22\cite{kaya2022neural} & BKC22\cite{kaya2022uncertainty} & BKW23*\cite{kaya2023multi}& PS-NeRF\cite{yang2022ps} & PS-Transformer\cite{pstransformer} & CasMVSNet\cite{gu2020cascade}-RT  & Ours\\
        \hline 
        BEAR & 0.504 & 0.987 & 0.926 & 0.895 & 0.965 & \textbf{0.994} & 0.496 & 0.902 & \underline{0.990}\\
        BUDDHA & 0.935 & 0.935 & 0.745 & 0.922 & \textbf{0.993} & \underline{0.970} & 0.387 & 0.913 & 0.953\\
        COW & 0.917 & 0.990 & 0.943 & 0.981 & \underline{0.987} & 0.984 & 0.617 & 0.896 & \textbf{0.993}\\
        POT2 & 0.459 & 0.985 & 0.929 & 0.909 & \underline{0.991} & 0.990 & 0.609 & 0.891 & \textbf{0.992}\\
        READING & 0.868 & 0.975 & 0.807 & 0.970 & 0.975 & 0.946 & 0.501 & \underline{0.981} & \textbf{0.989} \\
        \hline
        AVERAGE & 0.737 & 0.974 & 0.870 & 0.935 & \underline{0.982} & 0.977 & 0.522 & 0.917 & \textbf{0.983}\\
        \bottomrule
        \end{tabular}
        \end{center}
        \caption{F-score on L2 distance (higher is better) between reconstructed mesh and GT without ICP. `-RT' denotes trained on our synthetic MVPS dataset. For non-manual methods, the best result is shown in bold, 2nd best as underline. LZ20 \& PJ16 involve carefully crafted steps, manual efforts in finding correspondence, and an initial mesh or point cloud. BKW23* code not available, result from the paper.}
        \label{table:supp_3d_results_f_score}
        \end{table*}

\subsection{Comparison with TransMVSNet}
TransMVSNet \cite{transmvsnet} is a state-of-the-art MVS method leveraging a transformer to extract both intra-image global context and inter-image feature interaction. We retrain a TransMVSNet model on our synthetic MVPS dataset using hyper-parameters suggested by \cite{transmvsnet}. Each training sample has 3 views and test sample has 5 views, which is the same set-up as our method and the retrained CasMVSNet \cite{casmvsnet}.

We provide L1 Chamfer distance and F-score with L2 distance (threshold at 1mm) in Table \ref{table:supp_transmvsnet}, which demonstrate that our method outperform STOA generalizable MVS method. Also note that the transformer used in TransMVSNet increases the runtime of the method compared to CasMVSNet.

\begin{table*}
\footnotesize
\begin{center}
\begin{tabular}{P{4em}|P{4em}P{4.5em}P{5em}P{5em}c|P{4em}P{4.5em}P{5em}P{5em}c}
\toprule
Metrics & \multicolumn{5}{c|}{\textbf{L1 Chamfer distance}} & \multicolumn{5}{c}{\textbf{F-score (1mm)}}\\
\hline
Method & CasMVSNet \cite{casmvsnet} & CasMVSNet-RT & TransMVSNet \cite{transmvsnet} & TransMVSNet-RT  & Ours & CasMVSNet \cite{casmvsnet} & CasMVSNet-RT & TransMVSNet \cite{transmvsnet} & TransMVSNet-RT & Ours\\
\hline
BEAR & 2.00 & 1.47 & 1.02 & 1.48 & \textbf{0.80} & 0.789 & 0.911 & 0.962 & 0.882 & \textbf{0.991}\\
BUDDHA & 1.44 & 1.26 & 1.09 & 1.10 & \textbf{1.07} & 0.878 & 0.919 & 0.961 & 0.963 & \textbf{0.958}\\
COW & 2.73 & 1.27 & 1.15 & 1.05 & \textbf{0.77} & 0.658 & 0.914 & 0.927 & 0.941 & \textbf{0.993}\\
POT2 & 1.89 & 1.46& 1.10 & 1.05 & \textbf{0.82} & 0.799 & 0.901 & 0.956 & 0.964 & \textbf{0.994}\\
READING & 1.07 & 0.75 & 0.87 & 0.76 & \textbf{0.66} & 0.941 & 0.980 & 0.971& 0.978 & \textbf{0.988}\\
\hline
AVERAGE & 1.83 & 1.24 & 1.05 & 1.09 & \textbf{0.82} & 0.813 & 0.925 & 0.955 & 0.946 & \textbf{0.985}\\
\hline
Recon. Time/object & 22s & 22s & 52s & 52s & 105s & 22s & 22s & 52s & 52s &  105s\\
\bottomrule
\end{tabular}
\end{center}
\caption{Comparison with TransMVSNet on L1 Chamfer distance and F-score with L2 distance (threshold at 1mm) after ICP. CasMVSNet \cite{casmvsnet} and TransMVSNet \cite{transmvsnet} denote the pretrained models on DTU dataset \cite{dtu}. 'RT' denotes trained on our synthetic MVPS dataset. }
\label{table:supp_transmvsnet}
\end{table*}

\end{document}